\documentclass[11pt,a4paper]{article}
\usepackage{amssymb}
\usepackage{graphicx}
\usepackage{url}
\usepackage{CJKutf8}
\usepackage{amsmath}
\usepackage{algorithm}

\usepackage{multirow}
\usepackage{algorithmic}
\usepackage{tabularx}   
\usepackage{makecell}   
\usepackage{caption}
\usepackage{amsthm}
\usepackage{booktabs}
\usepackage[switch]{lineno}
\usepackage{fontawesome}  

\usepackage{authblk} 

\usepackage{graphicx}
\graphicspath{{images/}}

\begin{document}

\title{Revealing the structure-property relationships of copper alloys with FAGC}

\author[1]{Yuexing Han}
\author[1]{Ruijie Li}
\author[1]{Guanxin Wan}
\author[1]{Gan Hu}
\author[2,*]{Yi Liu\footnote{Email: yiliu@shu.edu.cn}} 
\author[1,3,*]{Bing Wang\footnote{Email: bingbignwang@shu.edu.cn}} 
\affil[1]{School of Computer Engineering and Science, Shanghai University, Shanghai 200444, China}
\affil[2]{Materials Genome Institute, Shanghai Engineering Research Center for Integrated Circuits and Advanced Display Materials, Shanghai University, Shanghai 200444, China}
\affil[3]{Key Laboratory of Silicate Cultural Relics Conservation (Shanghai University), Ministry of Education, Shanghai 200444, China}

\maketitle

\begin{abstract}
Cu-Cr-Zr alloys play a crucial role in electronic devices and the electric power industry, where their electrical conductivity and hardness are of great importance. However, due to the scarcity of available samples, there has been a lack of effective studies exploring the relationship between the microstructural images of Cu-Cr-Zr alloys and their key properties. In this paper, the FAGC feature augmentation method is employed to enhance the microstructural images of Cu-Cr-Zr alloys within a feature space known as the pre-shape space. Pseudo-labels are then constructed to expand the number of training samples. These features are then input into various machine learning models to construct performance prediction models for the alloy. Finally, we validate the impact of different machine learning methods and the number of augmented features on prediction accuracy through experiments. Experimental results demonstrate that our method achieves superior performance in predicting electrical conductivity (\(R^2=0.978\)) and hardness (\(R^2=0.998\)) when using the decision tree classifier with 100 augmented samples. Further analysis reveals that regions with reduced image noise, such as fewer grain or phase boundaries, exhibit higher contributions to electrical conductivity. These findings highlight the potential of the FAGC method in overcoming the challenges of limited image data in materials science, offering a powerful tool for establishing detailed and quantitative relationships between complex microstructures and material properties.
\end{abstract}


\section{Introduction}

Copper and its alloys have become indispensable engineering materials in various industrial fields. Among the different copper-based alloys, the Cu-Cr-Zr ternary alloy system exhibits remarkable comprehensive properties, establishing critical materials in  electronic devices and the electric power industry~\cite{li2022effects,yang2023recent}.
With the advancement of high-end manufacturing, the performance of existing Cu-Cr-Zr alloys can no longer fully meet the growing application demands~\cite{yang2022high,mao2022enhancing}. Exploring the relationship between the microstructure and material properties --- particularly hardness and electrical conductivity --- of the Cu-Cr-Zr ternary alloy system is therefore essential for improving its processing techniques. Electrical conductivity and hardness are the core indicators for evaluating the engineering performance of materials, directly influencing their service life and reliability.
 The addition of Cr can refine the grains, thereby enhancing the alloy  hardness; however, an excessive Cr content inevitably leads to a reduction in electrical conductivity~\cite{purcek2014effect,fu2017effect,xu2018effect}. In contrast, Zr plays a crucial role in promoting the refinement and spheroidization of Cr precipitates and forming Zr-rich phases at the grain boundaries, which strengthens the grain boundaries and improves the overall mechanical properties~\cite{zhang2017high}. Moreover, a small addition of Zr exerts only a minor influence on electrical conductivity~\cite{wang2022improved}.  As discussed above, there exists a close relationship between material microstructure and material properties, and uncovering the complex correlations between them has become one of the major challenges in materials science research. Substantial progress has been made in understanding the fundamental properties of the Cu-Cr-Zr ternary alloy system~\cite{ramprasad2017machine}. However, current studies still exhibit significant limitations in accurately predicting the electrical conductivity and hardness of these alloys based on microstructural images. In this work, we propose a novel machine learning approach that leverages microstructural images to precisely predict the electrical conductivity and hardness of Cu-Cr-Zr alloys.

Conventional approaches in materials research typically begin with the preparation of experimental samples, followed by measurements of various physical properties to characterize their different attributes. However, this research paradigm is highly dependent on experimental samples and requires large amounts of repetitive and complex work, resulting in relatively low efficiency. In recent years, with the rapid development of machine learning methods, there has been a growing trend toward designing machine learning models based on data analysis to address materials science problems. For example, Bhandari \textit{et al.}~\cite{bhandari2023machine} employed machine learning methods to predict the thermal conductivity of additively manufactured alloys based on composition, temperature data. Chen \textit{et al.}~\cite{chen2019machine} constructed a dataset and applied feature engineering combined with machine learning techniques to predict the lattice thermal conductivity of inorganic materials. Rohatgi \textit{et al.}~\cite{rohatgi2023application} utilized graphene processing data to predict the mechanical properties of copper-graphene composites. Liu \textit{et al.}~\cite{liu2022accelerated} developed a machine learning-guided high-throughput experimental strategy to accelerate the discovery of non-equimolar high-entropy alloys, significantly improving hardness prediction accuracy and enabling efficient identification of superhard compositions.

However, these traditional machine learning methods still face certain challenges, including difficulties in handling complex relationships between materials and performance, and limitations in extracting more information from images. In recent years, numerous studies have leveraged the exceptional representational capability of deep neural networks across diverse types of data to enable effective extraction of material microstructural information and microstructural images for performance prediction. For example, Han et al. utilized deep learning and digital image processing techniques to statistically analyze the precipitates, especially the size of the sigma phase, in SEM images of 2205 duplex stainless steel (DSS) \cite{han2023microstructural}. They reasonably estimated the average interfacial energy between the s phase and the g phase. The effective thermal conductivity of sintered silver is predicted with superior performance by convolutional neural networks (CNN) compared to that achieved by conventional machine learning regression models \cite{du2023highly}. Similarly, a model with an enhanced CNN structure is employed to predict the effective thermal conductivity of composites and porous media \cite{wei2018predicting}. The network model demonstrated strong performance on a dataset of 1,500 composites generated by a quadratic structure generating set \cite{wang2007mesoscopic}.
The work of Kondo \textit{et al.}~\cite{kondo2017microstructure}, who extracted microstructural features of ceramics using CNNs to predict ionic conductivity; Zhang \textit{et al.}~\cite{zhang2024machine}, who trained CNNs and combined the extracted features with machine learning methods to capture the relationship between microstructural characteristics and electrical conductivity in lithium-ion conducting oxide solid electrolytes; and Tu \textit{et al.}~\cite{tu2024microstructural}, who applied CNNs to extract morphological features from microstructural images of aluminum alloys and established correlations with their mechanical property.

The performance of convolutional neural networks is highly dependent on the scale of the training dataset. For small-scale datasets or specific material microstructural images, directly using raw images as inputs to neural network models often fails to achieve satisfactory performance. However, the  preparation methods of Cu-Cr-Zr alloys are inefficient and costly, making it difficult to achieve large-scale sample fabrication~\cite{guo2019microstructure,chen2018precipitation,jeyaprakash2023enhanced}. Moreover, process variations for a single alloy composition~\cite{ma2022microstructure,sun2020effects,xia2012study,batra2003precipitation} or one to two alloys with optimal performance are challenging to analyze in depth~\cite{wang2018retaining}. In such cases, the proper application of data augmentation techniques to enlarge the dataset can improve both predictive accuracy and generalization capability.

Data augmentation methods can be broadly categorized into three types. 
The first involves introducing perturbations to images through processing operations such as translation, rotation, cropping, or noise addition\cite{shorten2019survey}. These approaches are widely adopted for image augmentation in natural image domains but have seen limited application in materials microstructural images~\cite{guo2023data}.
The second approach leverages generative models, such as variational autoencoders (VAEs)\cite{kingma2013auto} or generative adversarial networks (GANs)\cite{goodfellow2014generative}, which learn from real data to generate new images. However, these generative models are highly data-dependent, and for small datasets—such as the Cu-Cr-Zr microstructural images used in this study—the quality of generated images is often unsatisfactory. The third approach focuses on augmenting features extracted from material compositions or material images. Because feature augmentation enhances the representational ability of data by adapting and changing the features, it does not depend on the data type. For example, the Simple Feature Augmentation (SFA) algorithm exploits the covariance information of each class to improve model performance\cite{li2021simple}, while the ``moment exchange'' technique proposed by Li \textit{et al.}~\cite{li2021feature} replaces current moment information with latent feature information to strengthen the feature representation. This feature-level augmentation is particularly suitable for small-sample scenarios.


The shape space theory, first introduced by Kendall~\cite{kendall1984shape}, is used to describe an object and all of its transformations in a non-Euclidean space. This theory has been applied to object recognition~\cite{han2013recognize,han2014recognizing} as well as data augmentation tasks~\cite{han2023fagc}. In the pre-shape space, the great-circle distance, also referred to as the Geodesic distance, is employed to measure the similarity between two sets of features. Between two sets of features, a Geodesic curve can be constructed between two samples in the pre-shape space to generate new feature data~\cite{evans2005curve,kenobi2010shape}. Building upon this framework, Han \textit{et al.}~\cite{han2023fagc} proposed the FAGC (Feature Augmentation on Geodesic Curve), which extracts image features through deep neural networks, projects them into the pre-shape space, and subsequently constructs a Geodesic curve to fit the sample distribution. New samples are then generated by sampling along the Geodesic curve, thereby achieving feature augmentation in the pre-shape space.

\begin{figure*}[htbp]
    \centering
    \includegraphics[width=\textwidth]{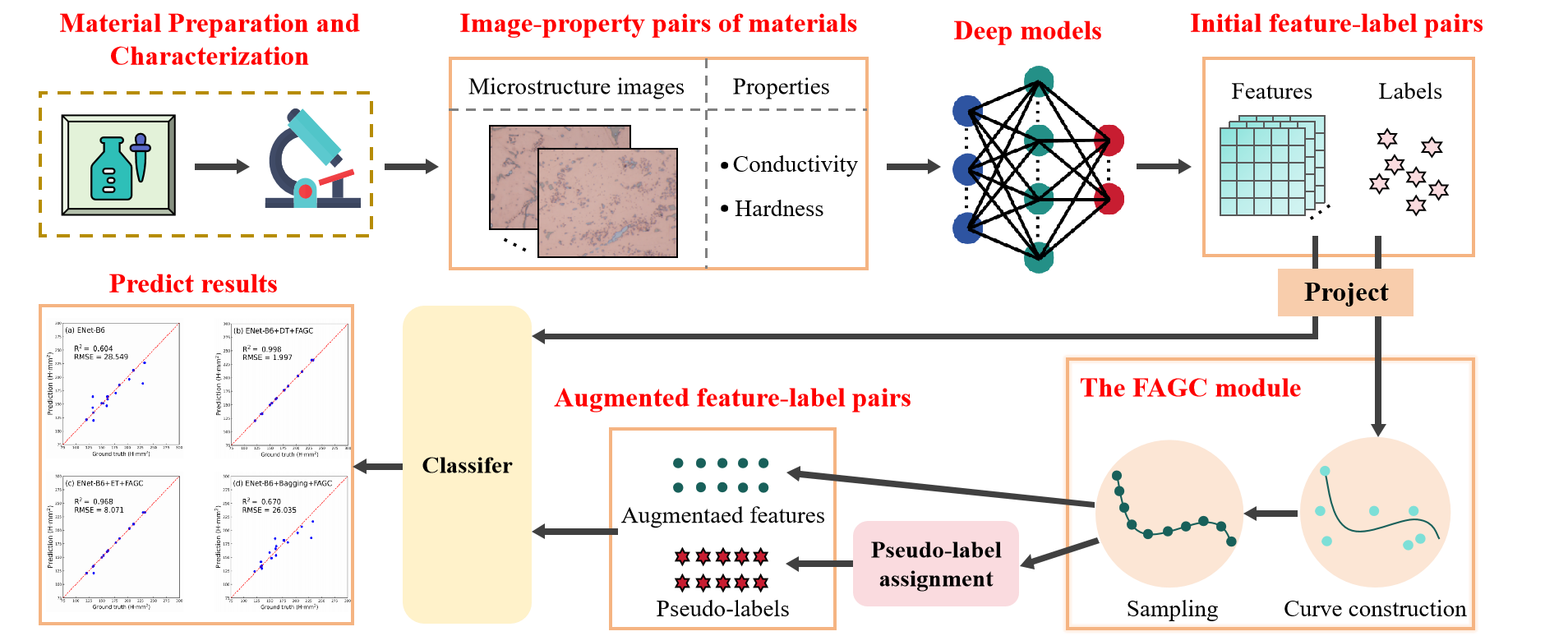}
    \caption{\rmfamily{An overview of performance prediction for electrical conductivity and hardness of Cu-Cr-Zr based on Geodesic curve feature augmentation.}}
    \label{fig:overview}
\end{figure*}

In this work, we investigate the structure–property relationships between the microstructure of Cu-Cr-Zr alloys and their mechanical and electrical properties, where the microstructural images are augmented using the FAGC method. 
Figure \ref{fig:overview} briefly summarizes the framework of Cu-Cr-Zr performance prediction: we first prepare and characterize a series of Cu-Cr-Zr alloys, acquire their microstructural images through scanning, and measure electrical conductivity and hardness as the target labels for performance prediction. Because of the small number of Cu-Cr-Zr alloy images, we employ the FAGC method to generate additional feature samples for data augmentation. Specifically, image features are extracted and projected into the pre-shape space, where new feature vectors are sampled along the Geodesic curve. A pseudo-labeling mechanism is then introduced to assign labels to the generated features, thereby expanding the feature–property dataset. By training regression models on the augmented dataset, where the extracted features serve as inputs, the prediction performance for material properties is significantly improved.

The results of this study provide systematic insights and predictive capability for interpreting the variations in alloy microstructures and product performance caused by changes in composition and processing parameters during industrial production. Furthermore, they offer valuable guidance for tailoring alloy properties and product quality through composition–process control, as well as for designing novel high-hardness and high-conductivity copper alloys.

\section{Methods}
\subsection{Material Preparation and Characterization}
The raw materials used in the experiments are pure Cu powder (purity $\geq$ 99.9\%, mass fraction, same below), pure Cr powder (purity $\geq$ 99.9\%), and a Cu-Zr intermediate alloy (purity $\geq$ 99.9\%). After alloy preparation via high-throughput (32-station) electrical arc melting, the macro Vickers hardness of the alloy samples was measured using a high-throughput (32-station) hardness testing system (RHT50-32, MTI Co. Ltd.). For each sample, at least five valid points were measured at different areas, and the average value was calculated. The electrical conductivity of the alloys was evaluated using an eddy current conductivity meter (FD102) compliant with industrial standards, with five valid measurements obtained for each alloy sample and averaged. Furthermore, an optical metallurgical microscope was employed to observe the microstructural images of the Cu-Cr-Zr alloys, including the morphology of the precipitate phases.

\subsection{Feature Extraction and Feature Augmentation}

\begin{figure*}[htbp]
    \centering
    \includegraphics[width=\textwidth]{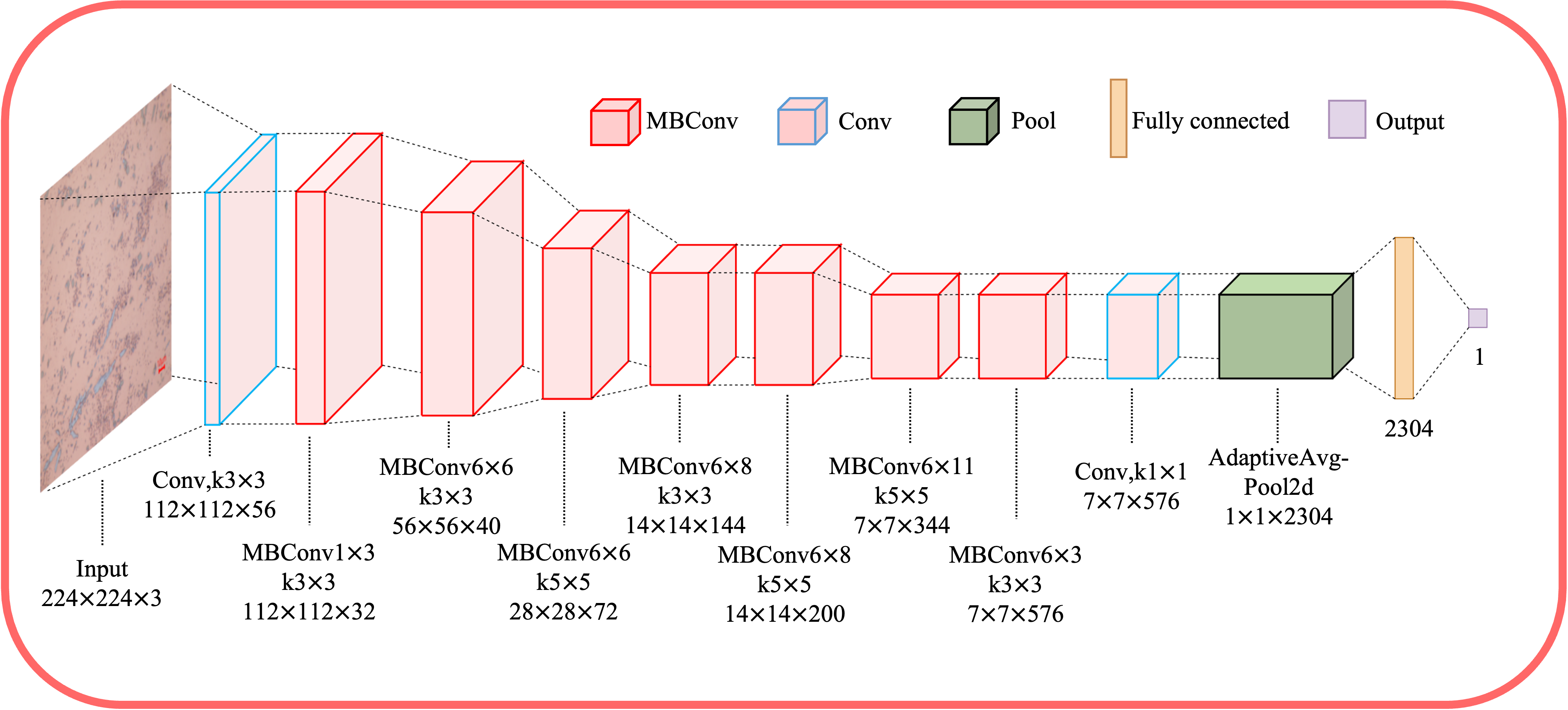}
    \caption{Architecture of EfficientNet-B6 for feature extraction. 
A 2304-dimensional feature vector is obtained after Global Average Pooling 
and mapped to a 1-dimensional regression output(conductivity or hardness).}
    \label{fig:efficientNet-b6}
\end{figure*}

Before applying the feature augmentation with the FAGC module, we extract the features from the input Cu-Cr-Zr images. Most neural network models can extract discriminative features, which improve the generalization ability of the model for processing the Cu-Cr-Zr data. EfficientNet-B6~\cite{tan2019efficientnet} is chosen as the feature extractor here and fine-tuned to fit our dataset. Because EfficientNet skillfully balances depth, width, and resolution through a unified scaling coefficient, it achieves superior image recognition performance while optimizing computational efficiency. As shown in Figure~\ref{fig:efficientNet-b6}, the architecture of EfficientNet-B6 consists of an input layer (i.e., material image), an output layer (i.e., material properties), and multiple convolutional layers.

In addition, the original EfficientNet-B6 model is a multi-class classifier, which is not directly applicable to the current task. Therefore, the output of the fully connected layers of the EfficientNet-B6 is adjusted to be compatible with the Cu-Cr-Zr image data. These modifications are crucial for predicting the performance metrics from Cu-Cr-Zr images. Meanwhile, the EfficientNet-B6 model has been pre-trained on the widely used ImageNet dataset~\cite{krizhevsky2012imagenet}, which contains about 1.2 million images. The training process for the feature extractor employs the mean square error as the loss function, and continues until convergence, where stabilization of the loss value indicates that the model has effectively learned and adapted to the training data. Then, the output data of the EfficientNet-B6 network is selected as the extracted feature data for the subsequent expansion of the features.

\begin{figure*}[htbp]
    \centering
    \includegraphics[width=\textwidth]{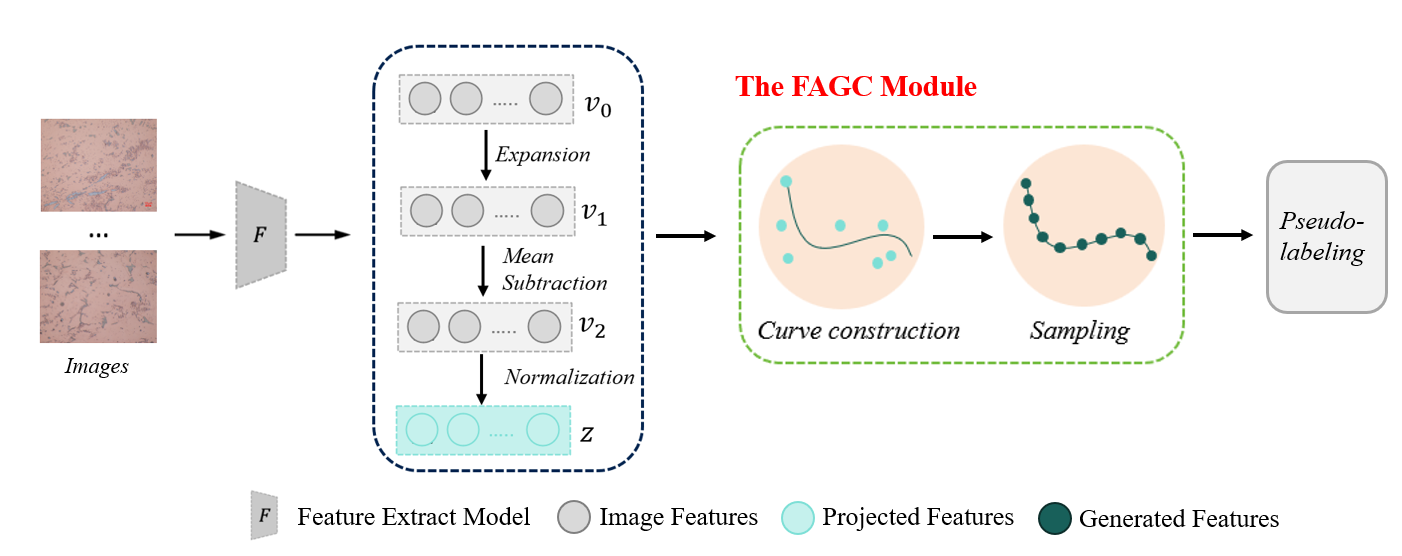}
    \caption{\rmfamily{The Feature Extraction and Feature Augmentation process. Cu-Cr-Zr images are projected into the pre-shape space, and FAGC process is performed to generate samples in the pre-shape space.}}
    \label{fig:shape-space}
\end{figure*}

We adopt the feature-level augmentation method (FAGC) proposed by Han \textit{et al.}~\cite{han2023fagc} to enhance the extracted features. Specifically, according to the shape space theory, the Cu-Cr-Zr image features obtained from the EfficientNet-B6 network need to be up-dimensioned before being projected into the pre-shape space. 
Here, we simply replicate each feature dimension to form pairs, i.e., a feature 
$v_0=\{x_1, x_2, \dots, x_n\}$ is transformed into 
$v_1=\{x_1, y_1, x_2, y_2, \dots, x_n, y_n \mid y_i = x_i, \, i \in [1,n]\}$. 
To remove positional bias, the mean value of each dimension is subtracted, resulting in $v_2$. 
Then, $v_2$ is normalized as $z = v_2 / \|v_2\|$ to project the feature into the pre-shape space. 
The projection process is illustrated in the left part of   Figure~\ref{fig:shape-space}.

The FAGC method identifies two representative feature vectors and constructs a Geodesic curve in the pre-shape space. A set of new features is then generated by uniformly sampling along this curve, forming the augmented feature $\{g_{1}, g_{2}, \dots, g_{K}\}$, as shown in the right part of Figure~\ref{fig:shape-space}.

\subsection{Pseudo-Labeling Strategy}

\begin{figure*}[htbp]
    \centering
    \includegraphics[width=\textwidth]{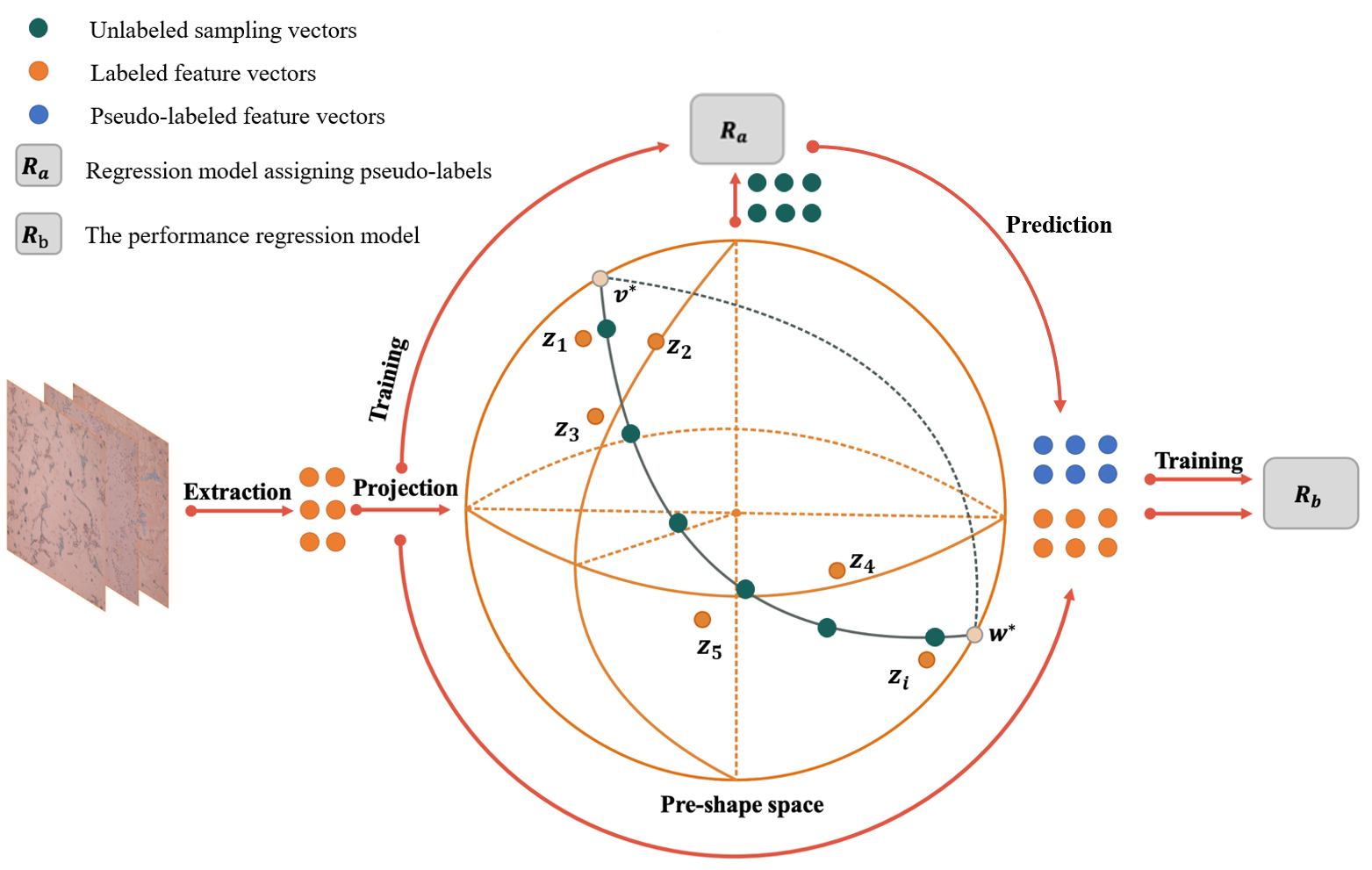}
    \caption{\rmfamily{The process of assigning pseudo-labels to generated features via FAGC in regression models.}}
    \label{fig:feature_augmentation}
\end{figure*}

In order to make the generated features applicable to downstream regression tasks, a pseudo-labeling mechanism is designed, as shown in Figure~\ref{fig:feature_augmentation}. 
The last fully connected layer of the EfficientNet-B6 network
is used as the regressor for the features, denoted as $R_a$. 
The generated features are then input into $R_a$, producing the predicted performance values
\(\tilde{y}_i = R_a(g_i), i=1,\dots,K \),
where $\tilde{y}_i $ is the corresponding predicted performance values. 
These predicted values serve as pseudo-labels for the generated feature set.

A new and enriched feature dataset is thus constructed by expanding the original feature set. 
The dataset is the combination of the original image feature set and the generated feature set with pseudo-labels, 
denoted as
\(Z' = \{(z_1, y_1), \dots, (z_M, y_M), (g_1, \tilde{y}_1), \dots, (g_K, \tilde{y}_K)\},\)
where $(z_i, y_i)$ are the original feature–label pairs and $(g_i, \tilde{y}_i)$ are the generated feature–pseudo-label pairs. 
Here, $M$ and $K$ are the numbers of the original and generated feature sets, respectively. 
Subsequently, other machine learning models are employed to train the expanded feature set $Z'$.

Since the pseudo-labels $\tilde{y}_i$ are produced by the regressor $R_a$, their quality inevitably influences downstream learning. 
To prevent label leakage, $R_a$ is trained using out-of-fold predictions to produce $\tilde{y}_i$. 
As further analyzed in Section~\ref{subsec:pseudo_label_quality}, regression performance consistently improves when the pseudo-labels are more accurate, demonstrating that FAGC-driven augmentation is beneficial as long as the teacher maintains reasonable predictive fidelity.

An optimal regression model $R_b$ is obtained through the training, which can predict the electrical conductivity and hardness properties of Cu-Cr-Zr more accurately. Figure~\ref{fig:feature_augmentation} illustrates the sequence of steps from selecting the generated feature set to training the final regression model.

\section{Results}
\subsection{Dataset and Experimental Settings}
\label{subsec:experiment settings}

\begin{figure}[htbp]
    \centering
    \begin{minipage}[b]{0.45\textwidth}
        \centering
        \includegraphics[width=\textwidth]{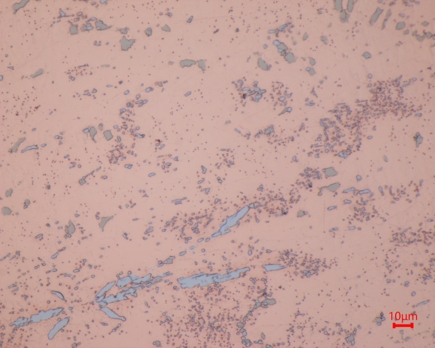}
        \vspace{2pt}
        \rmfamily{\textbf{(a)}
        Conductivity: 64.686~\%IACS;\\
        Hardness: 229.86~HV}
    \end{minipage}
    \hfill
    \begin{minipage}[b]{0.45\textwidth}
        \centering
        \includegraphics[width=\textwidth]{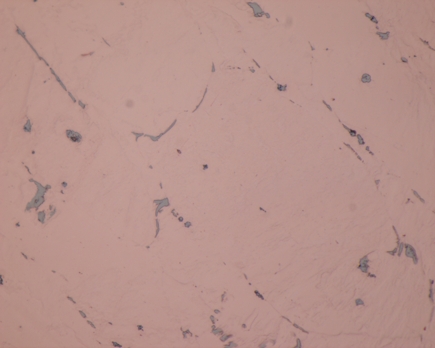}
        \vspace{2pt}
        \rmfamily{\textbf{(b)}
        Conductivity: 81.656~\%IACS;\\
        Hardness: 160.06~HV}
    \end{minipage}

    \vspace{6pt} 

    \begin{minipage}[b]{0.45\textwidth}
        \centering
        \includegraphics[width=\textwidth]{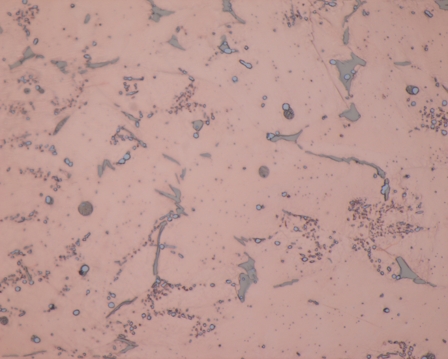}
        \vspace{2pt}
        \rmfamily{\textbf{(c)}
        Conductivity: 69.163~\%IACS;\\
        Hardness: 203.50~HV}
    \end{minipage}
    \hfill
    \begin{minipage}[b]{0.45\textwidth}
        \centering
        \includegraphics[width=\textwidth]{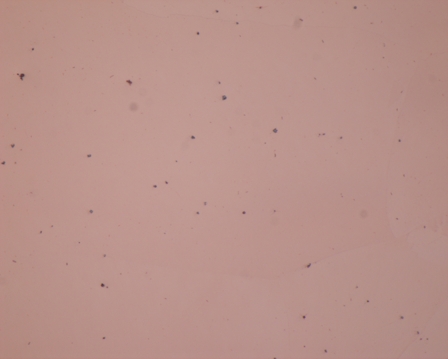}
        \vspace{2pt}
        \rmfamily{\textbf{(d)}
        Conductivity: 93.840~\%IACS;\\
        Hardness: 134.36~HV}
    \end{minipage}

    \vspace{6pt} 

    \begin{minipage}[b]{0.45\textwidth}
        \centering
        \includegraphics[width=\textwidth]{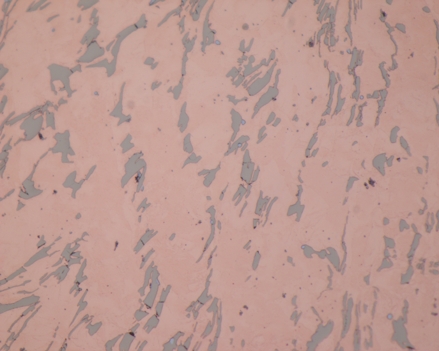}
        \vspace{2pt}
        \rmfamily{\textbf{(e)}
        Conductivity: 67.353~\%IACS;\\
        Hardness: 211.50~HV}
    \end{minipage}
    \hfill
    \begin{minipage}[b]{0.45\textwidth}
        \centering
        \includegraphics[width=\textwidth]{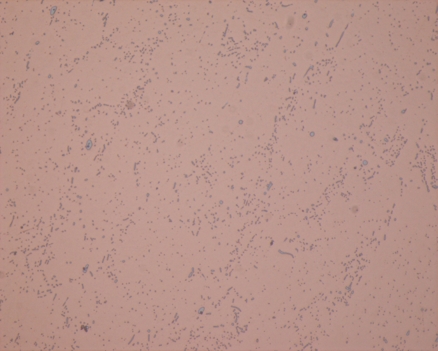}
        \vspace{2pt}
        \rmfamily{\textbf{(f)}
        Conductivity: 86.000~\%IACS;\\
        Hardness: 184.63~HV}
    \end{minipage}

    \caption{\rmfamily{Microstructural image samples of Cu-Cr-Zr alloys with corresponding electrical conductivity and hardness.}}
    \label{fig:Cu-Cr-Zr}
\end{figure}

\begin{figure}[htbp]
    \centering
    \begin{minipage}{0.45\textwidth}
        \centering
        \includegraphics[width=\textwidth]{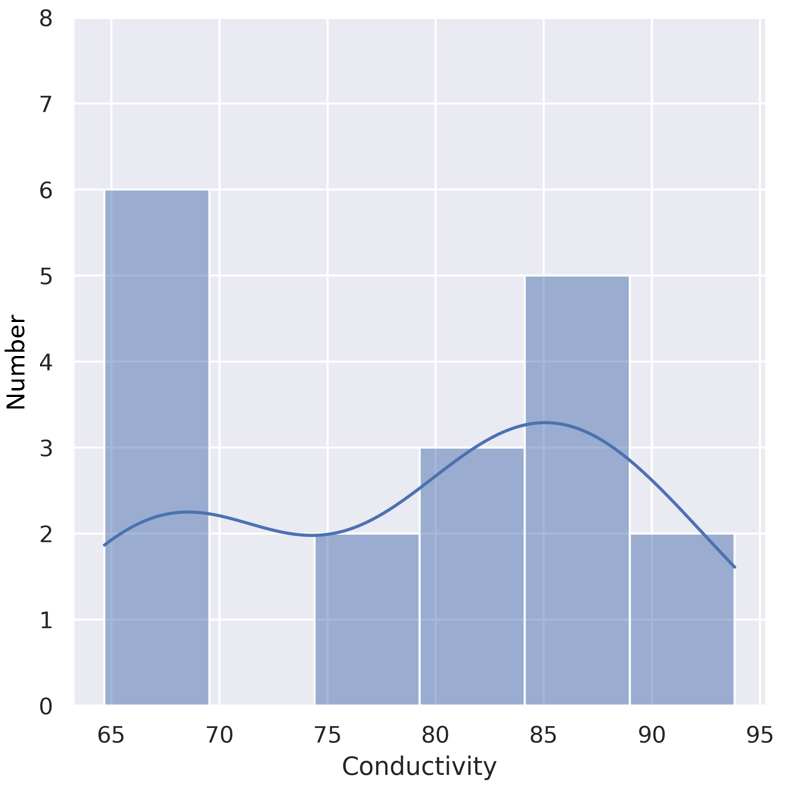}
    \end{minipage}
    \hfill
    \begin{minipage}{0.45\textwidth}
        \centering
        \includegraphics[width=\textwidth]{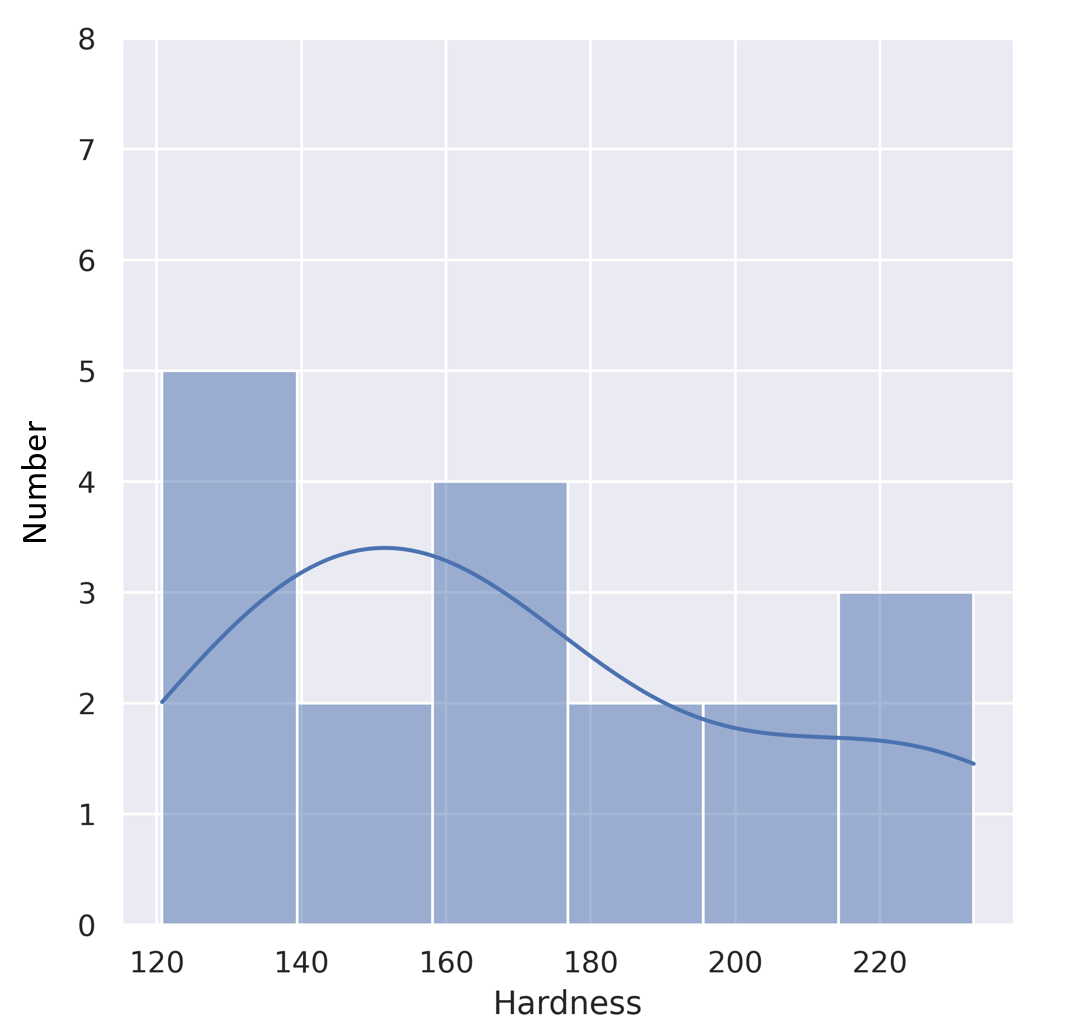}
    \end{minipage}
    \caption{\rmfamily{Experimentally collected distributions of electrical conductivity and hardness of Cu–Cr–Zr alloys. \textit{Number} indicates the number of samples within each property range.}}
    \label{fig:performance_distribution}
\end{figure}

Representative microstructural images of Cu–Cr–Zr alloys are shown in Figure~\ref{fig:Cu-Cr-Zr}, and Figure~\ref{fig:performance_distribution} illustrates the experimentally collected distributions of electrical conductivity and hardness of the alloys. The dataset obtained from our experiments consists of only 18 microstructural images, each paired with corresponding measurements of electrical conductivity and hardness. After partitioning the property ranges into 6 histogram bins, each bin contains merely 2--4 samples, indicating that the experimentally collected data exhibit pronounced sparsity and potential label imbalance. Such limited per-bin counts and small overall sample size pose significant challenges for training deep learning models on alloy microstructural images, leading to an increased risk of overfitting and reduced statistical reliability.


To determine the optimal EfficientNet model, EfficientNet-B0 through EfficientNet-B7 are trained. Here, the standard model performance metrics such as R$^2$-score, MAE, MSE, and RMSE are employed for performance evaluation.
The MAE (mean absolute error) is utilized to quantify the average absolute differences between predicted values and the actual observed values. The MSE (mean square error) and the RMSE (root mean square error) used as the indicators to describe the extent of deviation between predicted values and actual observations. 
The advantage of the RMSE is that it incorporates squared terms, thereby reducing the influence of outliers on model evaluation. Suppose the predicted values of the model are $\{y_{pred}^{1},y_{pred}^{2},... ,y_{pred}^{N}\}$, and the actual observations are $\{y_{true}^{1},y_{true}^{2},... ,y_{true}^{N}\}$, where $N$ is the number of image samples, then the metrics are calculated as follows:
\begin{equation}
    \text{R}^2=1-\frac{\sum_{i=1}^{N}(y_{true}^{i}-y_{pred}^{i})^2}{\sum_{i=1}^{N}(y_{true}^{i}-\bar{y}_{true})^2},
\end{equation}
and
\begin{equation}
    \text{MAE}=\frac{1}{N}\sum_{i=1}^{N}\Bigl| y_{true}^{i}-y_{pred}^{i} \Bigr|,
\end{equation}
and
\begin{equation}
    \text{RMSE}=\sqrt{\frac{1}{N}\sum_{i=1}^{N}(y_{true}^{i}-y_{pred}^{i})^2},
\end{equation}
where $\bar{y}_{true}$ denotes the average value of all observations.The value of R$^2$ takes on the range of $(-\infty,1]$. The closer the value is to 1, the better the generalization ability of the model. The values of MAE and RMSE take the range of $[0,+\infty)$. The closer the MAE value is to 0, the smaller the average prediction error of the model. The closer the RMSE value is to 0, the better the predictive performance of the model.


EfficientNet-B0 to EfficientNet-B7 are employed in a 6-fold cross-validation method to partition the dataset. In each iteration, a fold consists of 3 test images and 15 training images. The models are pre-trained on ImageNet to initialize the network weights. The loss function is set as the mean square error, and the Adam optimizer~\cite{kingma2014adam} is used with an initial learning rate of 0.0001. An adaptive learning rate decay strategy is applied over a total of 200 training epochs. The experimental results are presented in Table~\ref{tab:efficientNet}. As shown in the table, EfficientNet-B6 achieves better performance on most metrics. Therefore, EfficientNet-B6 is selected as the optimal feature extraction model based on the performance metrics.

\begin{table}[htbp]
    \rmfamily
    \centering
    \caption{\rmfamily{Comparative analysis of different EfficientNet(ENet) architectures for predicting conductivity and hardness in Cu-Cr-Zr alloy images.}}
    \begin{tabular}{ccccccc}
        \toprule
        \multirow{2}{*}{Network} & \multicolumn{3}{c}{Conductivity} & \multicolumn{3}{c}{Hardness}                                                                       \\ \cmidrule{2-7}
                                 & $R^2$                            & MAE                          & RMSE           & $R^2$          & MAE             & RMSE            \\ \midrule
        ENet-B0                  & 0.326                            & 9.674                        & 10.348         & 0.389          & 32.532          & 35.428          \\
        ENet-B1                  & 0.334                            & 9.452                        & 10.287         & 0.427          & 31.975          & 34.317          \\
        ENet-B2                  & 0.487                            & 8.375                        & 9.031          & 0.496          & 29.863          & 32.191          \\
        ENet-B3                  & 0.559                            & 8.164                        & 8.376          & 0.559          & 28.613          & 30.079          \\
        ENet-B4                  & 0.556                            & 7.963                        & 8.395          & 0.553          & 27.462          & 30.291          \\
        ENet-B5                  & 0.570                            & 7.682                        & 8.263          & 0.557          & 27.521          & 30.172          \\
        ENet-B6                  & \textbf{0.606}                   & 7.415                        & \textbf{7.918} & \textbf{0.619} & \textbf{26.883} & \textbf{27.981} \\
        ENet-B7                  & 0.591                            & \textbf{6.872}               & 8.061          & 0.598          & 28.947          & 28.763          \\
        \bottomrule
    \end{tabular}
    \label{tab:efficientNet}
\end{table}

\subsection{Performance Prediction of Cu-Cr-Zr Alloys}

To comprehensively evaluate the influence of the FAGC module on regression models, two sets of experiments are conducted to predict the conductivity and hardness of Cu-Cr-Zr. The machine learning methods include Linear Regression (LR), KNN Regressor (KNN)~\cite{cover1967nearest}, AdaBoost~\cite{freund1996experiments}, ExtraTree (ET)~\cite{geurts2006extremely}, DecisionTree (DT)~\cite{breiman2017classification}, and Bagging~\cite{breiman1996bagging}, either used directly or combined with the FAGC module.

\begin{figure}[htbp]
    \centering
    \begin{minipage}[b]{0.45\textwidth}
        \centering
        \includegraphics[width=\textwidth]{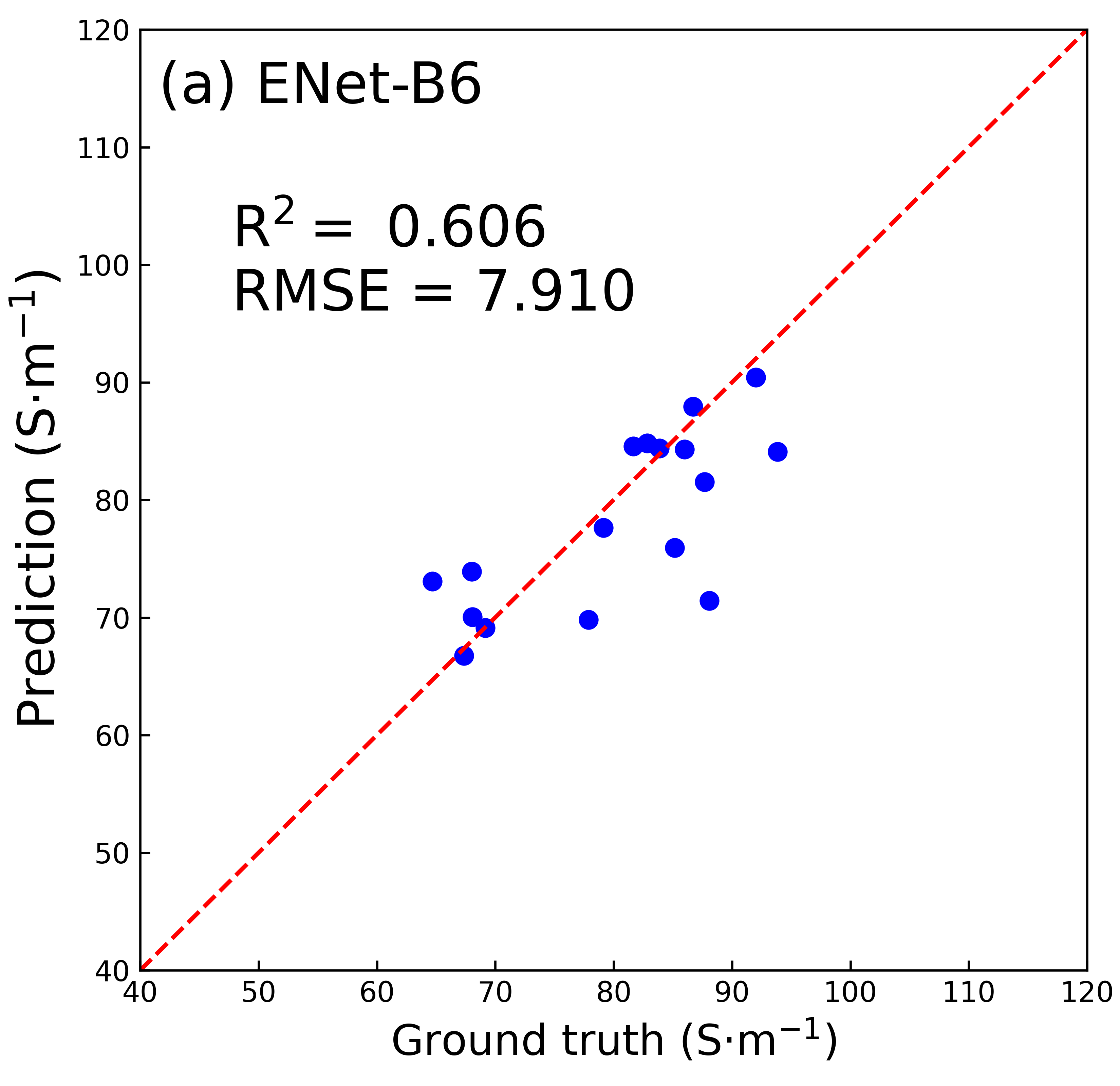}
        \label{fig:conductivity-1}
    \end{minipage}
    \hfill
    \begin{minipage}[b]{0.45\textwidth}
        \centering
        \includegraphics[width=\textwidth]{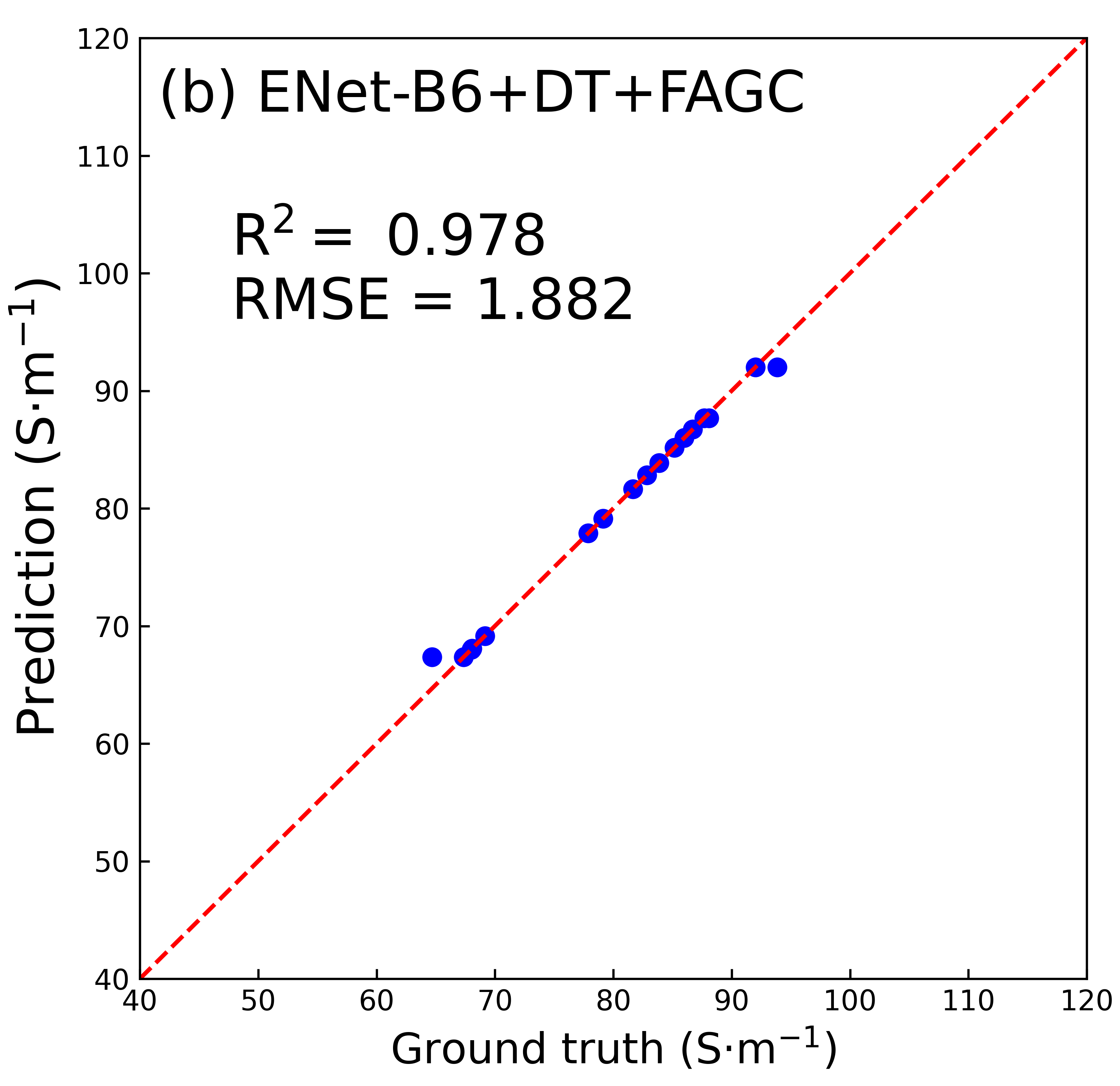}
        \label{fig:conductivity-2}
    \end{minipage}
    \hfill
    \begin{minipage}[b]{0.45\textwidth}
        \centering
        \includegraphics[width=\textwidth]{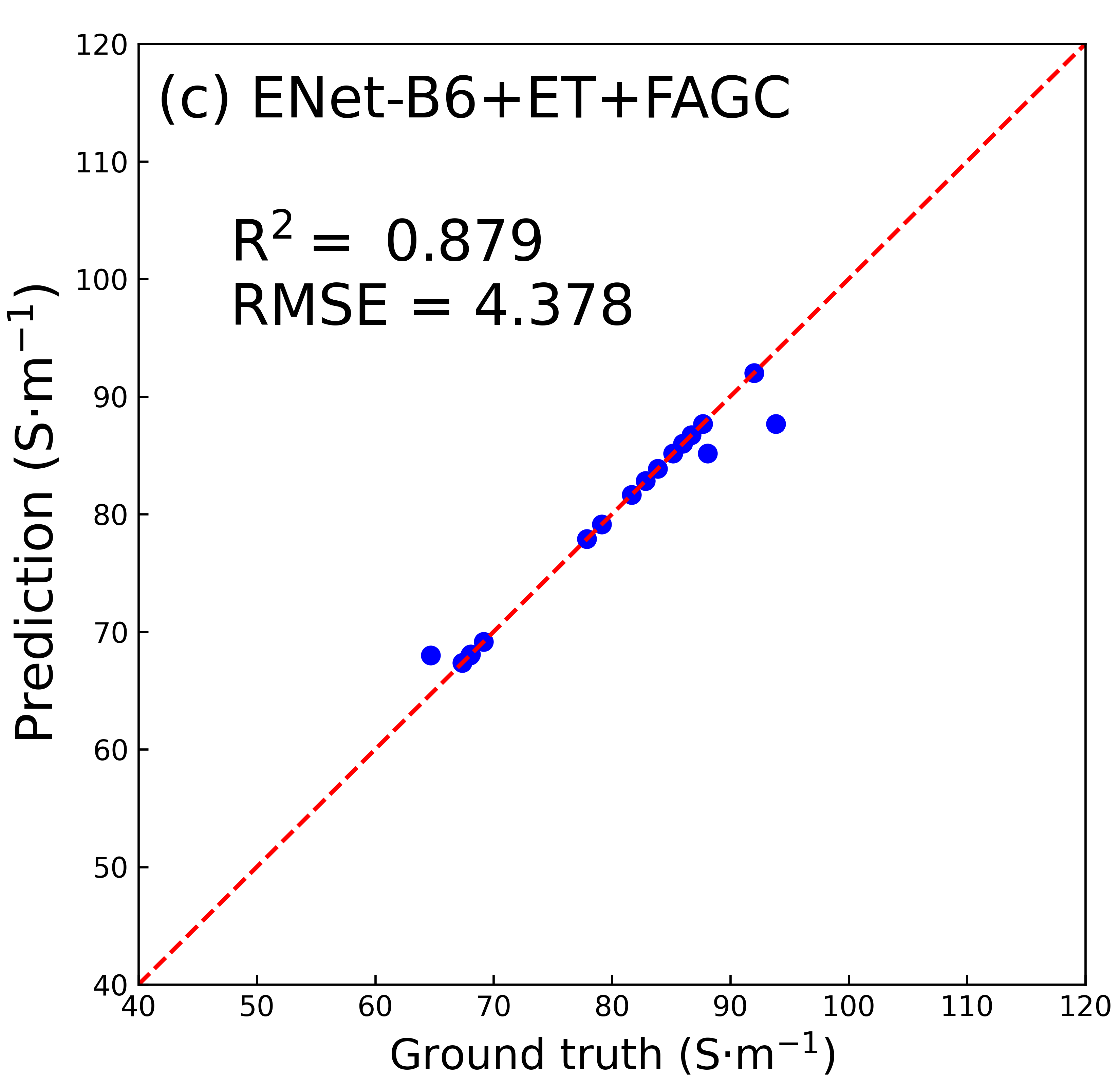}
        \label{fig:conductivity-3}
    \end{minipage}
    \hfill
    \begin{minipage}[b]{0.45\textwidth}
        \centering
        \includegraphics[width=\textwidth]{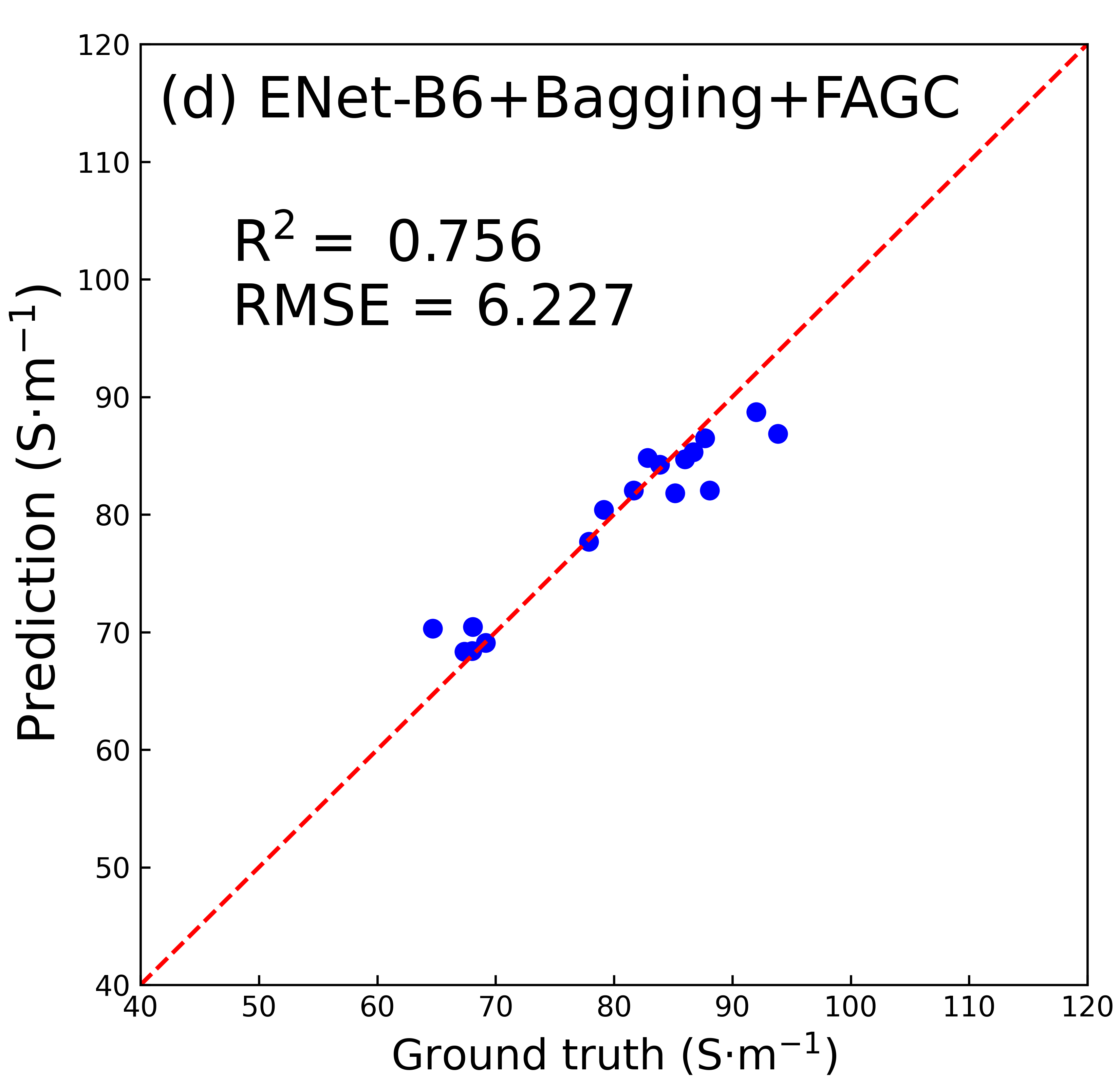}
        \label{fig:conductivity-4}
    \end{minipage}
    \caption{\rmfamily{Comparison of the prediction and observation of electrical conductivity in Cu-Cr-Zr with FAGC across different regression models.}}
    \label{fig:conductivity-FAGC-ablation}
\end{figure}

Figure \ref{fig:conductivity-FAGC-ablation} illustrates the prediction results for Cu-Cr-Zr conductivity. The red dashed line represents a perfect correlation between the actual and predicted values. Each blue point corresponds to a two-dimensional coordinate pair of predicted and actual values. The closer a blue point is to the red dashed line, the closer the prediction is to the actual Cu-Cr-Zr conductivity. The experimental results demonstrate the significant influence of the FAGC module in improving regression performance. The performance of machine learning models is significantly enhanced after incorporating the FAGC module. In particular, the DecisionTree (DT) model achieves an R$^2$ of 0.978, with RMSE of 1.882. In addition, the experimental results illustrate the generalization and effectiveness of the FAGC module across different regression models. More experiment results are shown in Table \ref{tab:conductivity}.

\begin{table}[htbp]
    \centering
    \rmfamily
    \caption{\rmfamily{Performance prediction of electrical conductivity in Cu-Cr-Zr using the EfficientNet-B6(ENet-B6) network with FAGC.}}
    \begin{tabular}{lcccc}
        \toprule
        Regressor                                    & $R^2$          & MAE            & RMSE           \\
        \midrule
        LR                                           & 0.548          & 8.136          & 8.476          \\
        KNN \cite{cover1967nearest}                  & 0.605          & 7.423          & 7.915          \\
        AdaBoost\cite{freund1996experiments}         & 0.663          & 6.957          & 7.319          \\
        ExtraTree\cite{geurts2006extremely}          & 0.763          & 4.910          & 6.132          \\
        DecisionTree\cite{breiman2017classification} & 0.609          & 7.192          & 7.883          \\
        Bagging\cite{breiman1996bagging}             & 0.689          & 6.829          & 7.026          \\
        ENet-B6                                      & 0.606          & 7.415          & 7.918          \\
        \midrule
        ENet-B6 + LR + \textbf{FAGC}                 & 0.510          & 8.740          & 8.825          \\
        ENet-B6 + KNN + \textbf{FAGC}                & 0.660          & 7.021          & 7.352          \\
        ENet-B6 + AdaBoost + \textbf{FAGC}           & 0.710          & 6.505          & 6.785          \\
        ENet-B6 + ExtraTree + \textbf{FAGC}          & 0.879          & 4.134          & 4.378          \\
        ENet-B6 + DecisionTree + \textbf{FAGC}       & \textbf{0.978} & \textbf{1.632} & \textbf{1.882} \\
        ENet-B6 + Bagging + \textbf{FAGC}            & 0.756          & 6.202          & 6.227          \\
        \midrule
    \end{tabular}
    \label{tab:conductivity}
\end{table}

\begin{figure}[htbp]
    \centering
    \begin{minipage}[b]{0.45\textwidth}
        \centering
        \includegraphics[width=\textwidth]{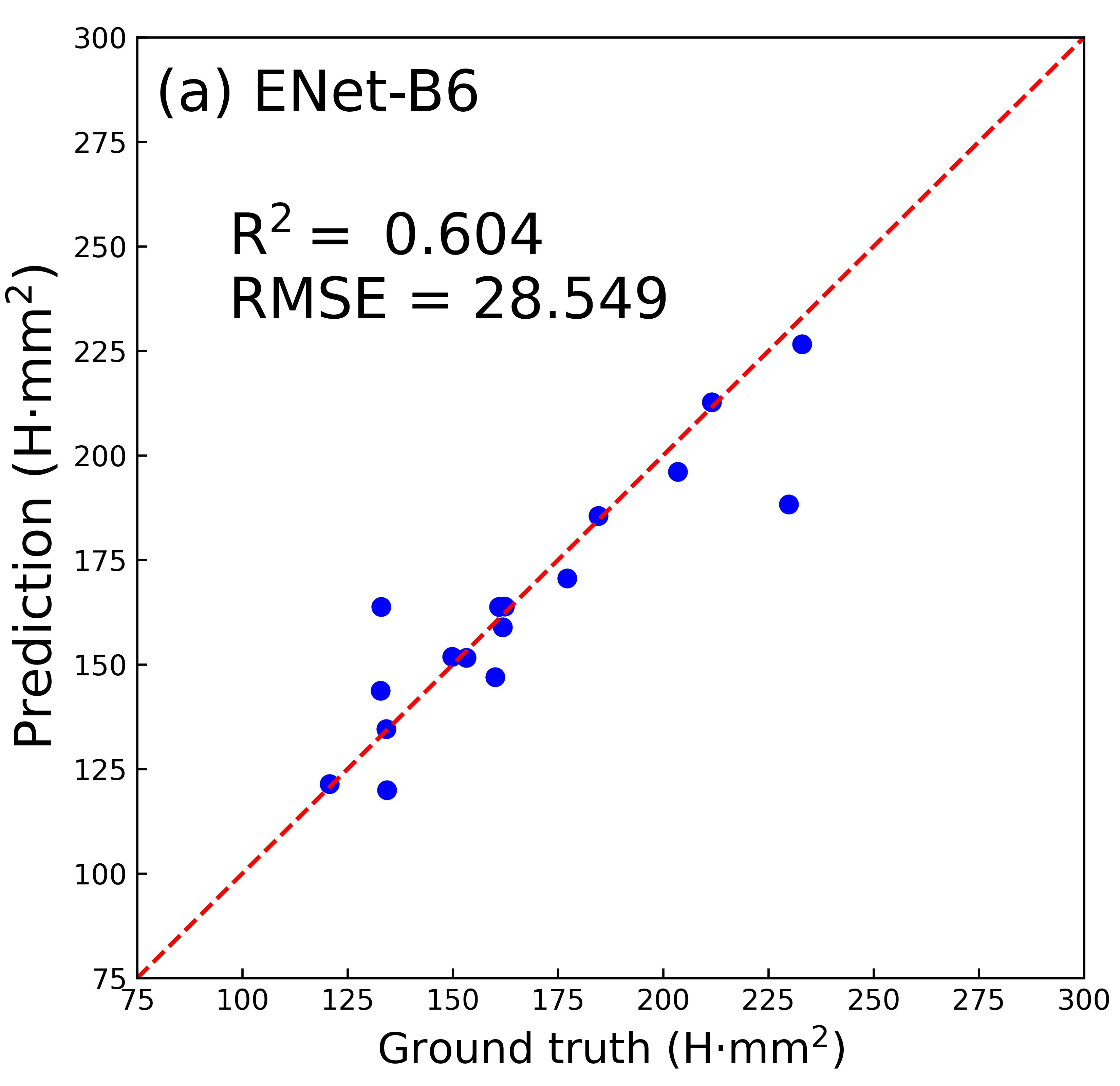}
        \label{fig:hardness-1}
    \end{minipage}
    \hfill
    \begin{minipage}[b]{0.45\textwidth}
        \centering
        \includegraphics[width=\textwidth]{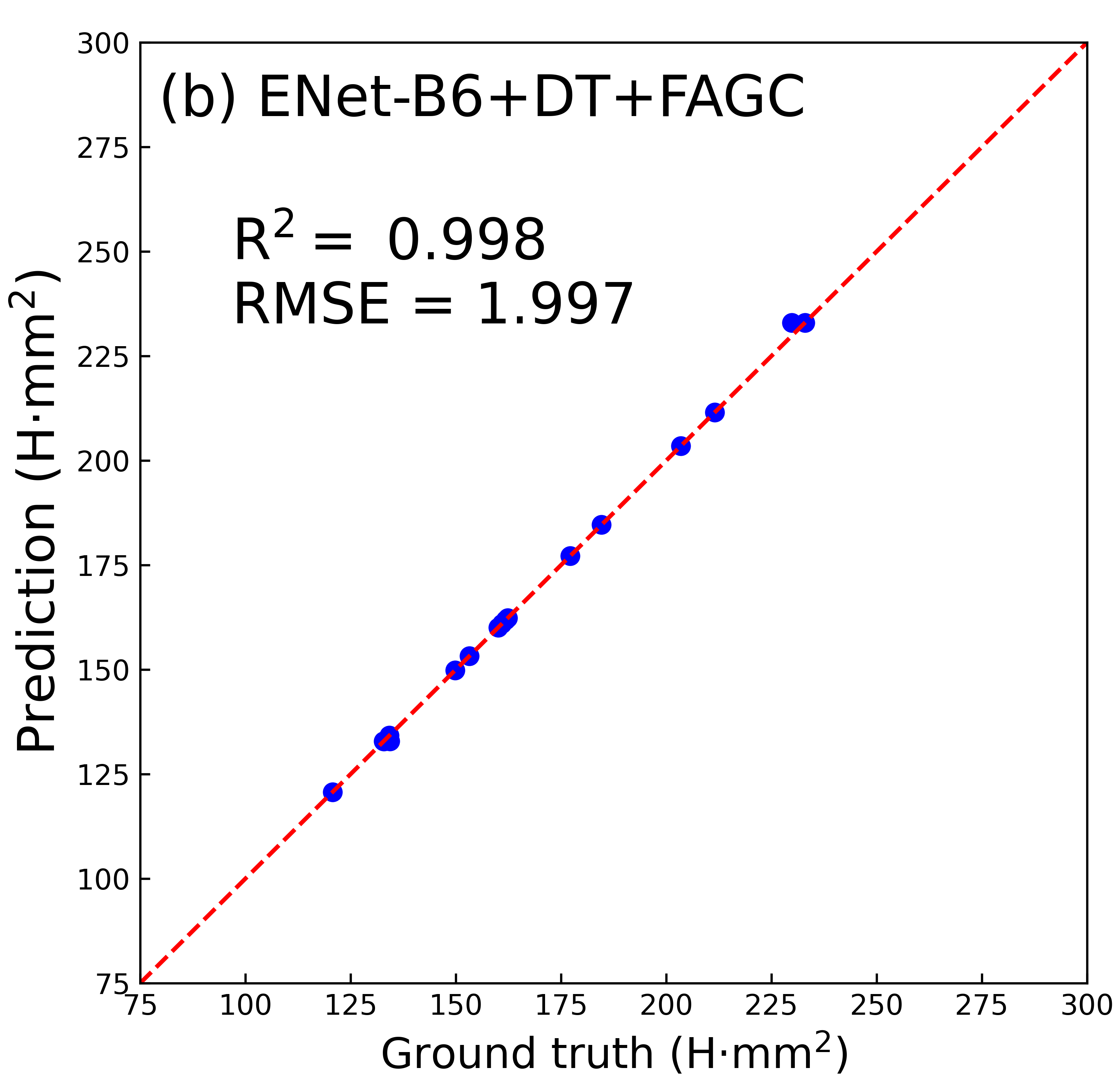}
        \label{fig:hardness-2}
    \end{minipage}
    \hfill
    \begin{minipage}[b]{0.45\textwidth}
        \centering
        \includegraphics[width=\textwidth]{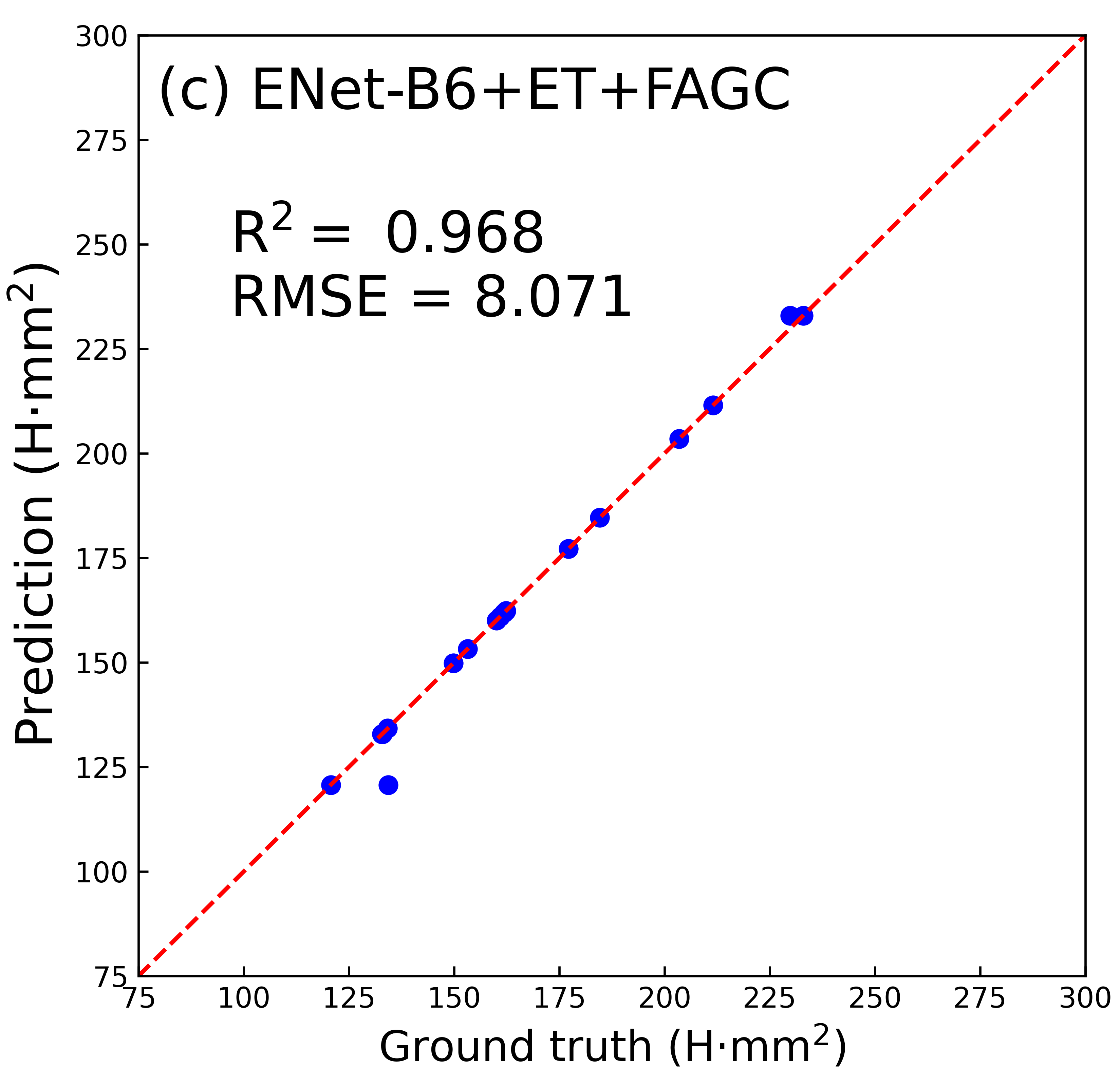}
        \label{fig:hardness-3}
    \end{minipage}
    \hfill
    \begin{minipage}[b]{0.45\textwidth}
        \centering
        \includegraphics[width=\textwidth]{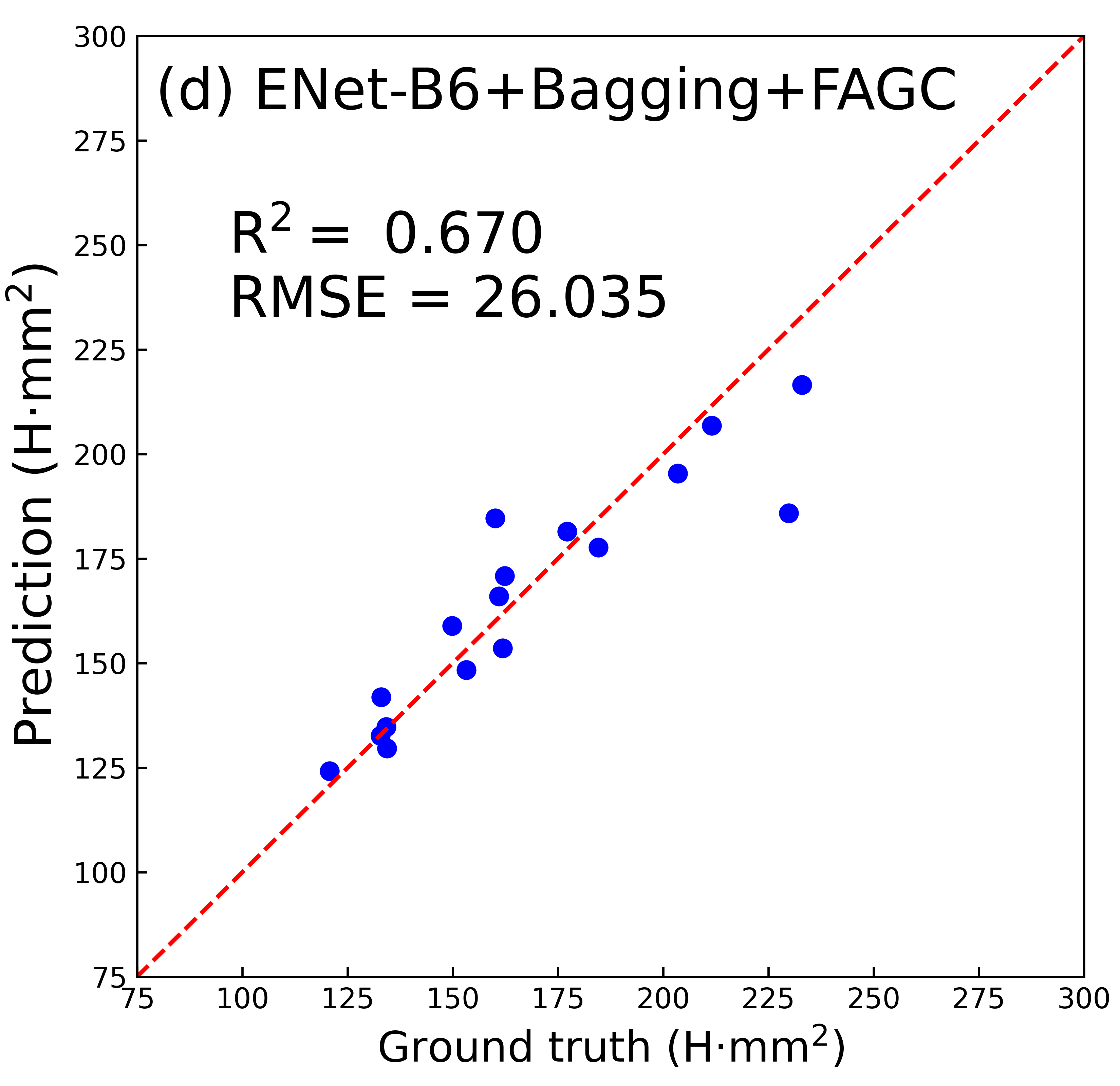}
        \label{fig:hardness-4}
    \end{minipage}
    \caption{\rmfamily{Comparison of the prediction and observation of hardness in Cu-Cr-Zr with FAGC across different regression models.}}
    \label{fig:hardness-FAGC-ablation}
\end{figure}

For the hardness performance prediction of Cu-Cr-Zr, a range of machine learning regression models are employed and compared with the FAGC module, following the experimental approaches detailed in Section~\ref{subsec:experiment settings}.   Figure \ref{fig:hardness-FAGC-ablation} illustrates a comparative analysis of the prediction results. For the hardness prediction of Cu-Cr-Zr, the EfficientNet-B6 network reaches an R$^2$ of 0.604, and an RMSE of 28.549. The introduction of the FAGC module significantly improves the predictive accuracy of these regression models across all metrics. The improved prediction accuracy is attributed to the effectiveness of the Geodesic curve in fitting the feature distribution region, even with a limited set of feature data. The effective utilization of feature data from Geodesic curves and the expansion of the dataset are achieved through the pseudo-labeling mechanism, which contributes to the enrichment of dataset diversity. Especially for the DecisionTree and ExtraTree models, the R$^2$ values reach 0.998 and 0.968, respectively. The experimental results once again demonstrate the potential effectiveness and versatility of the FAGC module in predicting the hardness performance of Cu-Cr-Zr.
More experiment results are shown in Table \ref{tab:hardness}. The DecisionTree model notably outperforms others in predicting hardness.

\begin{table}[htbp]
    \centering
    \rmfamily
    \caption{\rmfamily{Performance prediction of hardness in Cu-Cr-Zr using the EfficientNet-B6(ENet-B6) network with FAGC.}}
    \begin{tabular}{lcccc}
        \toprule
        Regressor                                    & $R^2$          & MAE            & RMSE           \\
        \midrule
        LR                                           & 0.483          & 28.632         & 32.558         \\
        KNN \cite{cover1967nearest}                  & 0.539          & 26.891         & 30.785         \\
        AdaBoost\cite{freund1996experiments}         & 0.504          & 23.207         & 31.918         \\
        ExtraTree\cite{geurts2006extremely}          & 0.550          & 18.118         & 30.416         \\
        DecisionTree\cite{breiman2017classification} & 0.638          & 19.675         & 27.272         \\
        Bagging\cite{breiman1996bagging}             & 0.430          & 24.971         & 34.226         \\
        ENet-B6                                      & 0.604          & 26.883         & 28.549         \\
        \midrule
        ENet-B6 + LR  + \textbf{FAGC}                & 0.477          & 28.779         & 32.780         \\
        ENet-B6 + KNN + \textbf{FAGC}                & 0.557          & 25.970         & 30.193         \\
        ENet-B6 + AdaBoost + \textbf{FAGC}           & 0.915          & 11.057         & 13.219         \\
        ENet-B6 + ExtraTree + \textbf{FAGC}          & 0.968          & 5.632          & 8.071          \\
        ENet-B6 + DecisionTree + \textbf{FAGC}       & \textbf{0.998} & \textbf{1.598} & \textbf{1.997} \\
        ENet-B6 + Bagging + \textbf{FAGC}            & 0.670          & 19.188         & 26.035         \\
        \hline
    \end{tabular}
    \label{tab:hardness}
\end{table}

\subsection{t-SNE Visualization of Generated and Actual Features}
\begin{figure}[htbp]
    \centering
    \includegraphics[width=0.8\linewidth]{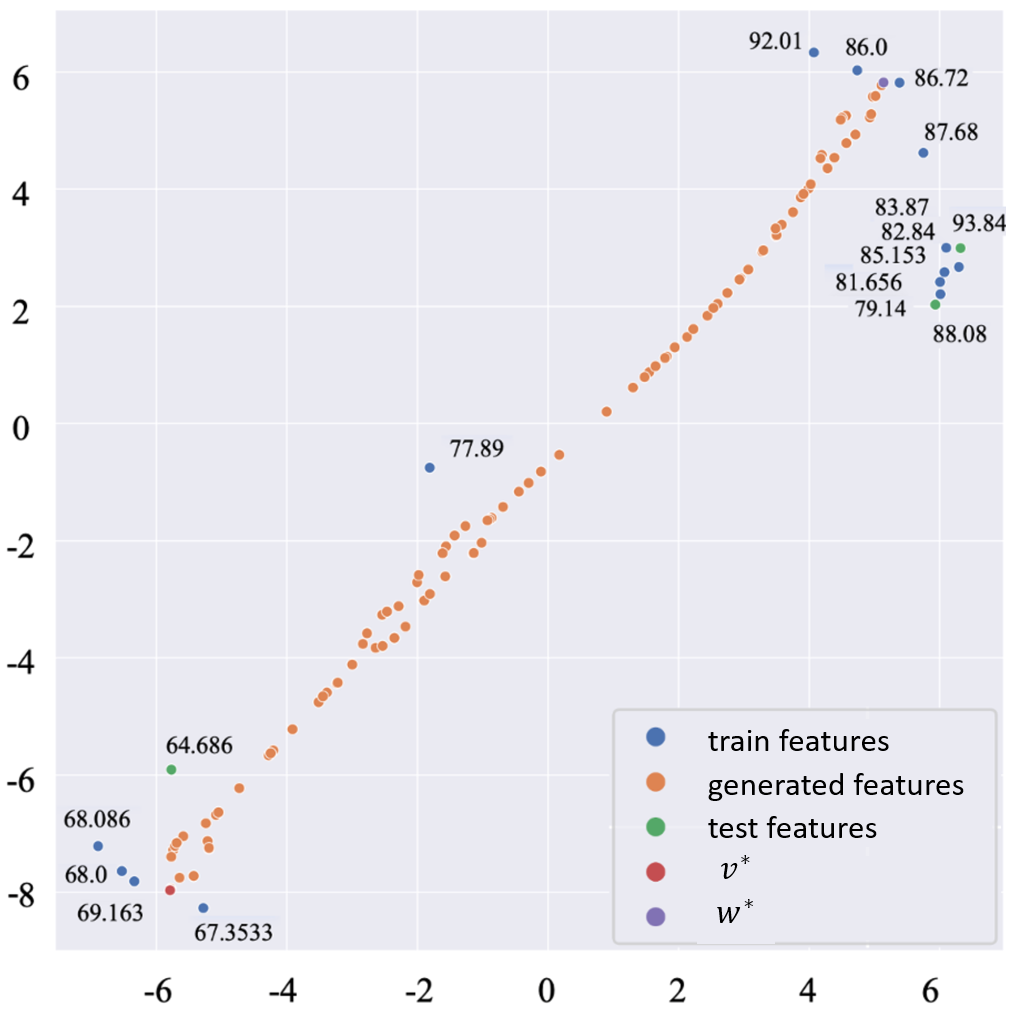}
    \caption{\rmfamily{Visualization for the generated features and training features of the Cu-Cr-Zr alloys through t-SNE.}}
    \label{fig:t-sne}
\end{figure}

To investigate the correlation between the generated features and the actual features of the Cu-Cr-Zr material, we use the t-SNE (t-distributed stochastic neighbor embedding) method to reduce the dimensionality of the feature data for visualization. By mapping the feature data into a two-dimensional space, we observe the distribution of the generated and actual features of the Cu-Cr-Zr material, as shown in Figure \ref{fig:t-sne}. In Figure \ref{fig:t-sne}, the training and test features of Cu-Cr-Zr are represented by blue and green dots, respectively. The orange dots represent the generated features, while the red and purple dots denote the two endpoints of the optimal Geodesic curve. The Geodesic curve can fit the feature region of Cu-Cr-Zr, thereby verifying the effectiveness of the FAGC module. The conductivity property values of Cu-Cr-Zr are labeled in the figure. The characteristics of Cu-Cr-Zr are mainly distributed in the lower left and upper right regions, where the conductivity performance values exhibit an increasing trend from the lower left to the upper right. It is worth noting that the trend of conductivity performance is strongly correlated with the distribution of the generated features, indicating that assigning pseudo-labels to the generated features is reasonable and effective.

\subsection{Effect of the Number of Generated Features}

\begin{figure}[htbp]
    \centering
    \includegraphics[width=\linewidth]{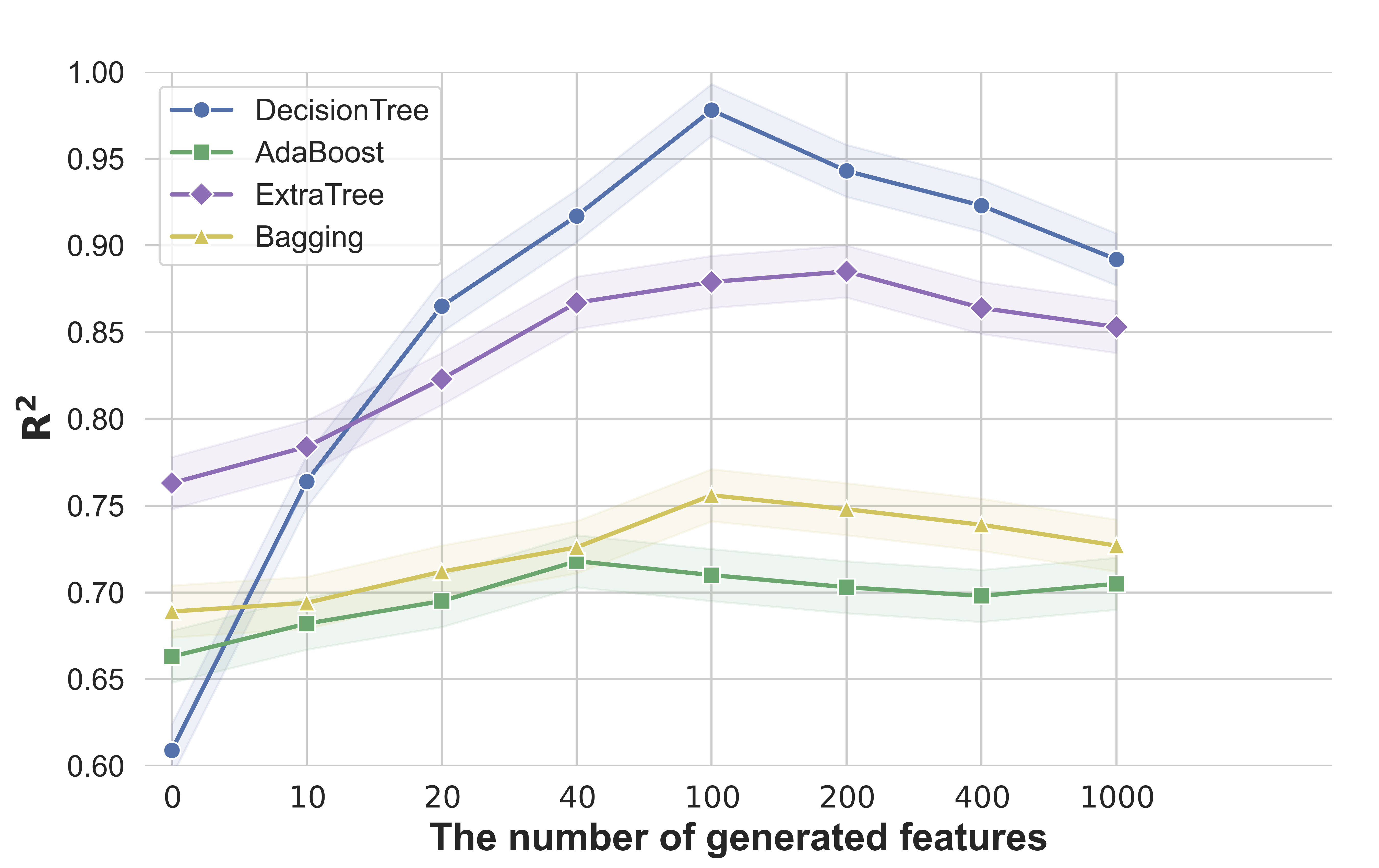}
    \caption{\rmfamily{The R$^2$ of the different numbers of the generated feature vectors in Cu-Cr-Zr conductivity prediction.}}
    \label{fig:generated-quantity-conductivity}
\end{figure}

\begin{figure}[htbp]
    \centering
    \includegraphics[width=\linewidth]{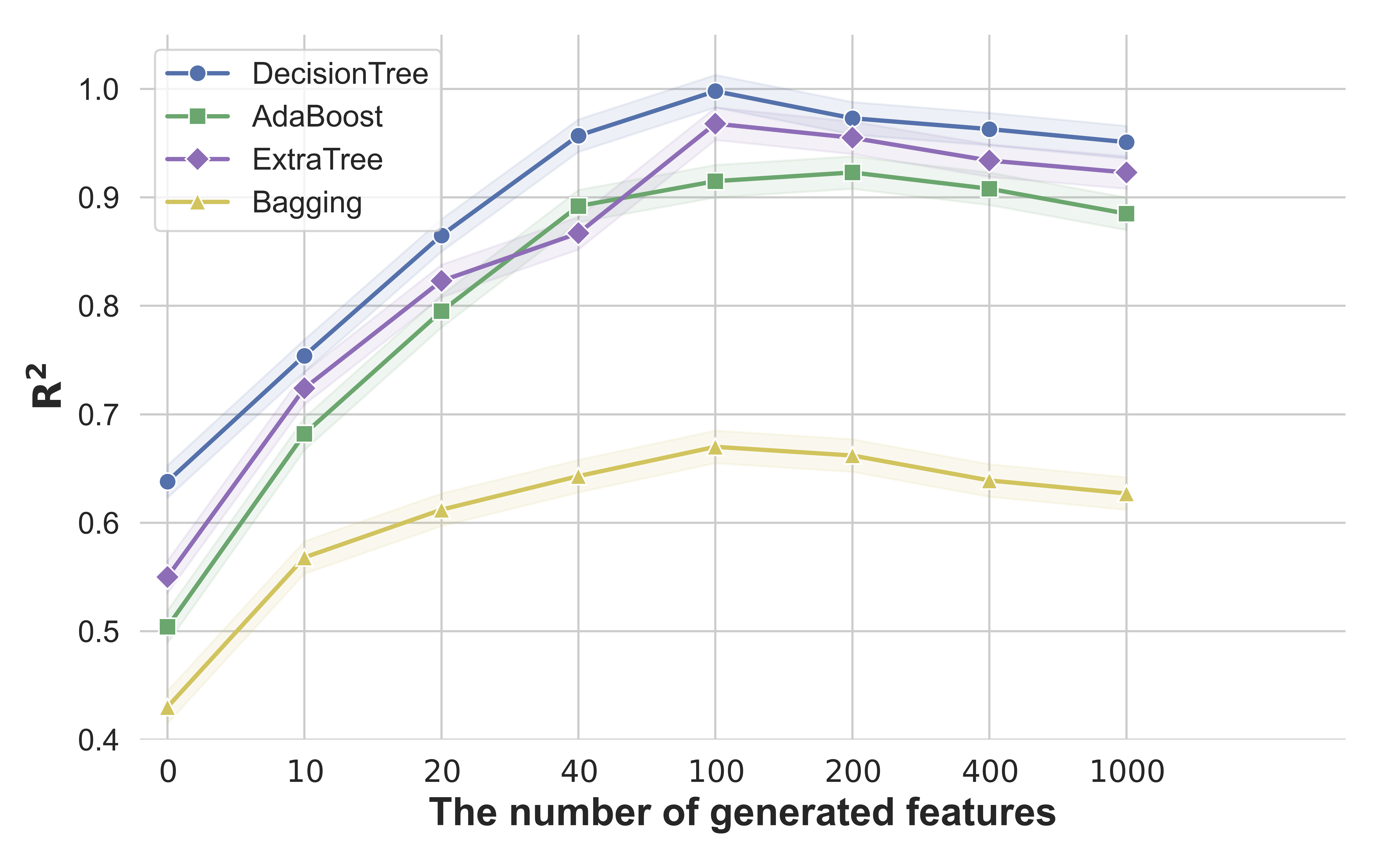}
    \caption{\rmfamily{The R$^2$ of the different numbers of the generated feature vectors in Cu-Cr-Zr hardness prediction.}}
    \label{fig:generated-quantity-hardness}
\end{figure}

An experiment is conducted to evaluate the influence of the number of features generated from the FAGC module. Several machine learning regression models are used to predict the performance of Cu-Cr-Zr conductivity and hardness, including AdaBoost, ExtraTree, DecisionTree, and Bagging. The feature extraction model employed is EfficientNet-B6, with the number of generated features including 10, 20, 40, 100, 200, 400, and 1000. The experimental results are presented in Figure \ref{fig:generated-quantity-conductivity} and Figure \ref{fig:generated-quantity-hardness}. The experimental results indicate that increasing the number of features generated can significantly improve the predictive ability of the model. However, the number of generated features and $R^2$ of the regression model is not strictly monotonous. Excessive generated features can cause the model to focus too much on the distribution of generated features, neglecting the distribution of authentic features. Therefore, the performance of the model does not grow in proportion to the size of the generated features. In our Cu-Cr-Zr dataset, about 100 features are generated to obtain the optimal results, which is about five times the size of the original dataset. Under these conditions, most models demonstrate superior predictive performance.

\subsection{Effect of Pseudo-Label Quality}
\label{subsec:pseudo_label_quality}
\begin{figure*}[htbp]
    \includegraphics[width=\textwidth]{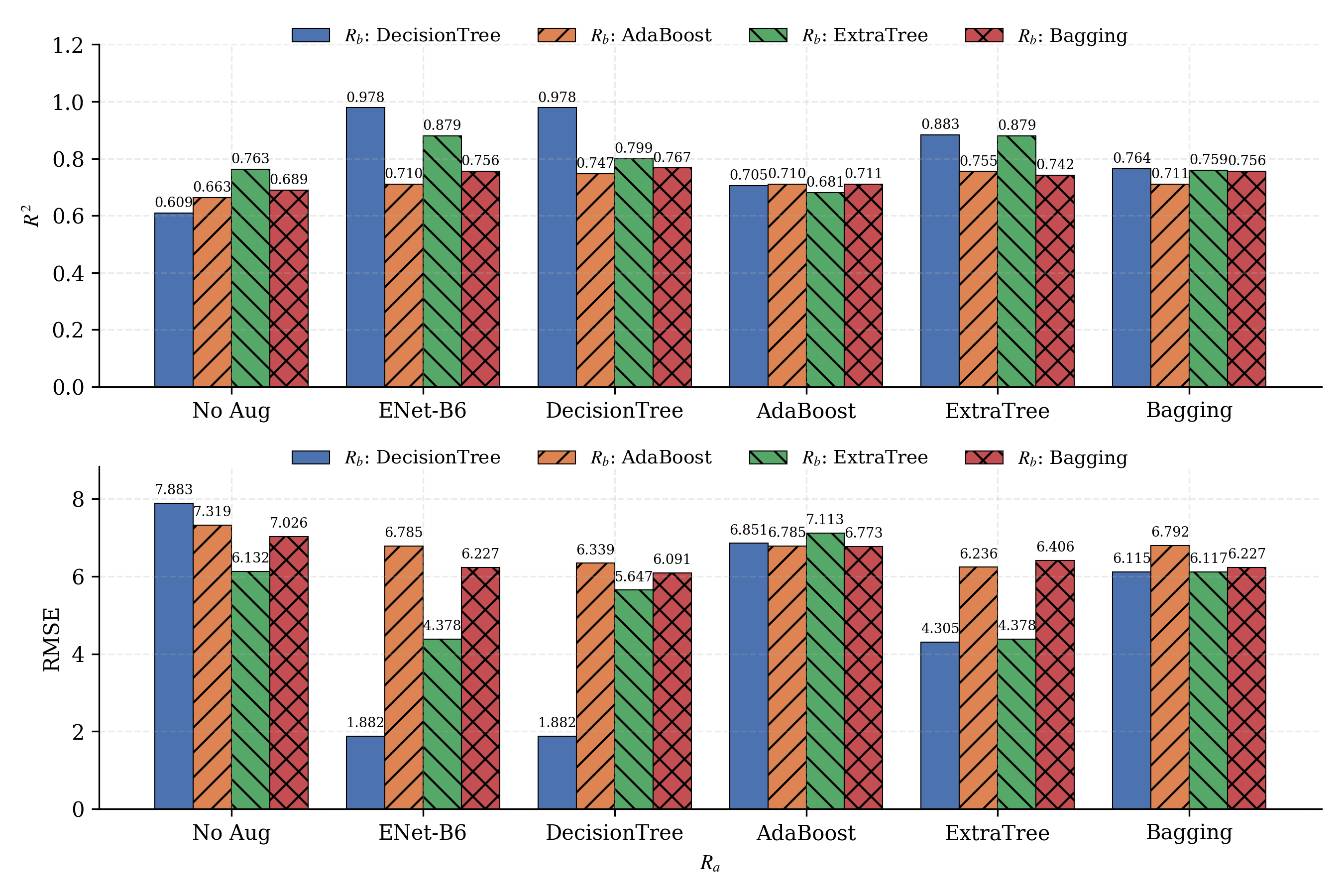}
    \caption{\rmfamily{The influence of pseudo-label quality on regression models in Cu-Cr-Zr conductivity prediction. $R_a$ means the regression model for predicting pseudo-labels. $R_b$ represents the conductivity prediction model.}}
    \label{fig:regressor-pseudo-conductivity}
\end{figure*}

\begin{figure*}[htbp]
    \includegraphics[width=\textwidth]{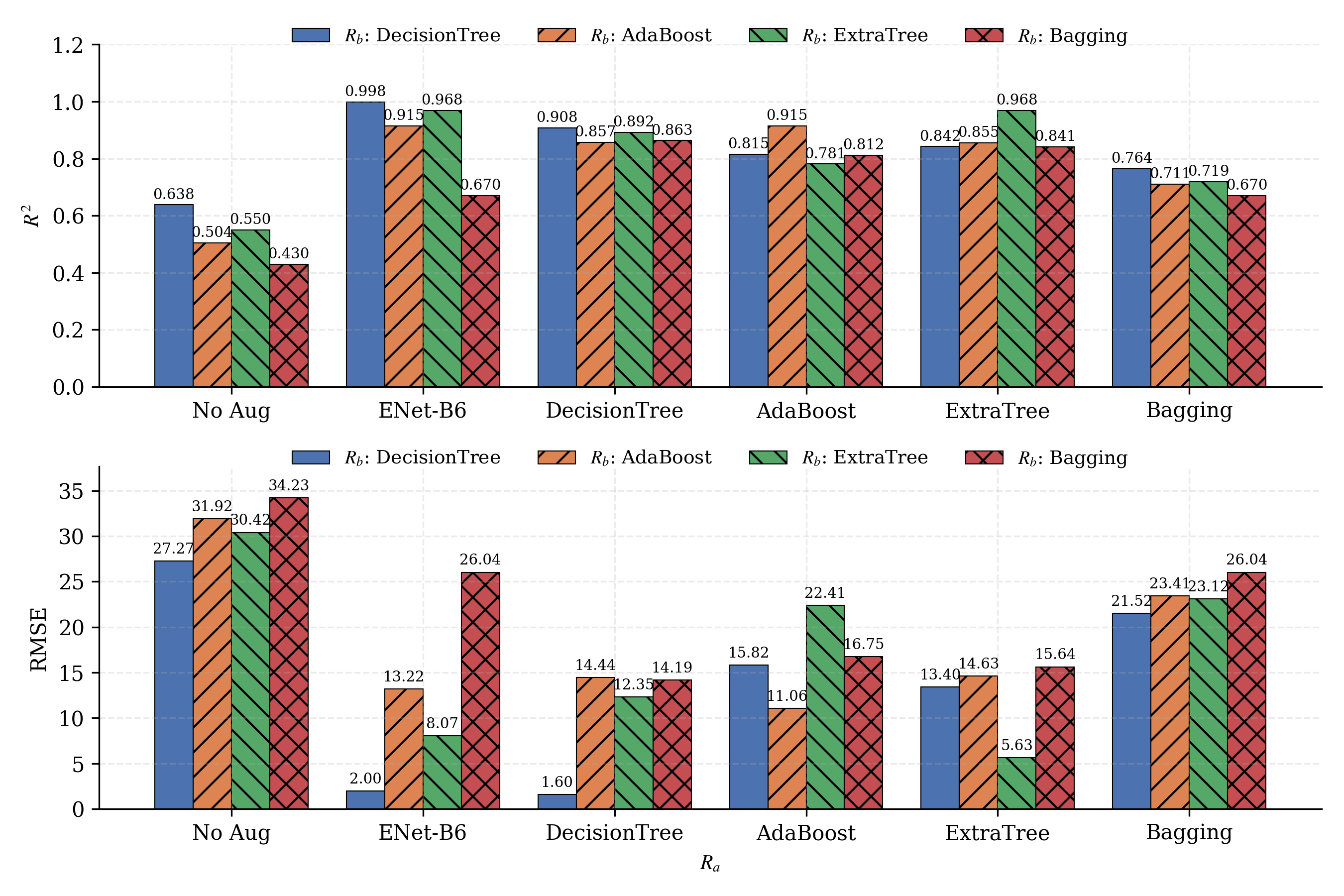}
    \caption{\rmfamily{The influence of pseudo-label quality on regression models in Cu-Cr-Zr hardness prediction. $R_a$ means the regression model for predicting pseudo-labeling. $R_b$ represents the hardness prediction model.}}
    \label{fig:regressor-pseudo-hardness}
\end{figure*}

In order to evaluate the influence of pseudo-label quality on the performance of the regression models, a series of experiments were conducted to predict the electrical conductivity and hardness of Cu-Cr-Zr alloys and to evaluate pseudo-labels generated by EfficientNet-B6, DecisionTree, AdaBoost, ExtraTree, and Bagging. The results of the comparative experiments are illustrated in Figure~\ref{fig:regressor-pseudo-conductivity} and Figure~\ref{fig:regressor-pseudo-hardness}, and the regression model training strategy follows the procedure described in Section~\ref{subsec:experiment settings}. 
For example, in Figure~\ref{fig:regressor-pseudo-conductivity}, the upper panel shows the bar chart of the coefficient of determination ($R^2$). The unpatterned blue bar in the ``No Aug'' column, labeled as 0.609, indicates that when no feature augmentation is applied, using DecisionTree as $R_b$ results in an $R^2$ value of 0.609 for the conductivity regression task. In the lower panel, which illustrates the RMSE values, green bar with right-leaning diagonal hatching in the ``ENet-B6'' column represents the case where FAGC-based feature augmentation is employed, ENet-B6 serves as the pseudo-label regression model $R_a$, and ExtraTree is used as the conductivity prediction model $R_b$, yielding an RMSE of 4.378. The experimental results show that the quality of the pseudo-labels has a significant influence on the performance of the regression models. Both EfficientNet-B6 and the DecisionTree models achieve good performance in terms of R$^2$ and RMSE.

\subsection{Application to Materials Analysis}
\begin{figure}[htbp]
    \centering
    \includegraphics[width=\linewidth]{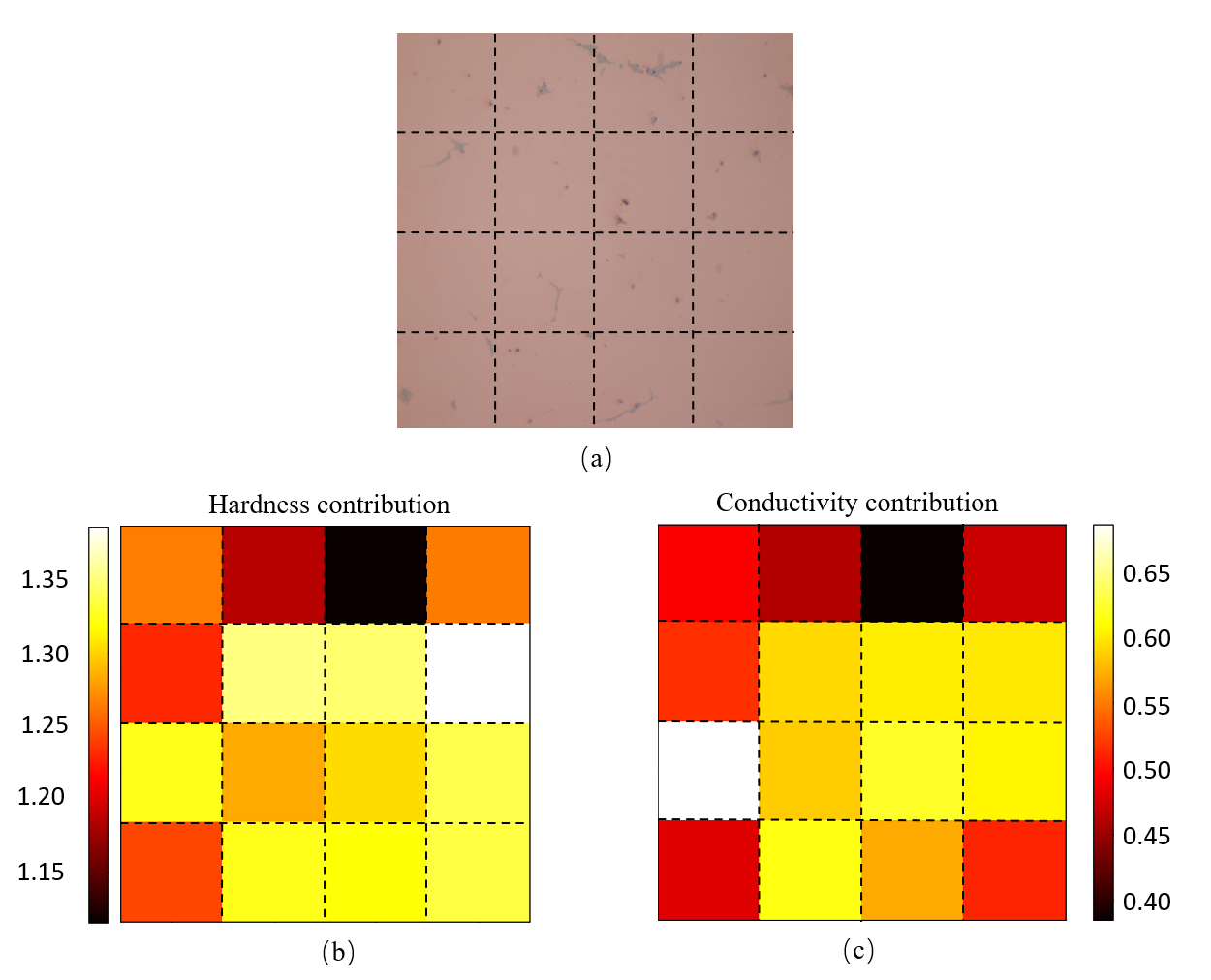}
    \caption{\rmfamily{
(a) Microstructural image of copper alloy; 
(b) prediction of the contribution of different image regions on the hardness performance; 
(c) prediction of the contribution of different image regions on the conductivity performance.}}
    \label{fig:apply}
\end{figure}

After establishing the structure–property relationship model for Cu alloys, further analysis of local information within material images can provide insights for the development of high-performance new materials. The copper alloy microstructural images were partitioned into 16 equal segments and fed into the trained performance prediction model, followed by visualization of the results. The visualized results are presented in Figure \ref{fig:apply}. 
Under ideal conditions, a homogeneous microstructure corresponds to stable and smooth spatial distributions of both conductivity and hardness. Lighter-colored regions in Figure~\ref{fig:apply}(b) and (c) indicate higher values of the corresponding property, while darker-colored areas in Figure~\ref{fig:apply}(b) and (c) represent lower values.

Figure~\ref{fig:apply}(b) illustrates the influence of different microstructural regions on hardness performance. Regions exhibiting denser or more complex microstructures—such as finer grains or more precipitates—tend to be associated with increased hardness, which is reflected as lighter-colored zones on the heatmap. Conversely, areas with coarser grains or fewer  precipitates typically show lower hardness values, manifesting as darker regions. These spatial variations highlight the strong dependence of mechanical properties on local microstructural features.

In contrast, Figure~\ref{fig:apply}(c) visualizes the spatial distribution of conductivity. Under ideal circumstances, a homogeneous microstructure would correspond to a stable and smooth conductivity distribution, where lighter colors indicate higher conductivity and darker colors represent lower conductivity. As seen in the microstructural image (a), regions with fewer grain boundaries or fewer  precipitates generally correspond to higher conductivity (lighter zones in (c)), whereas regions with complex grain boundary networks or localized inhomogeneities tend to exhibit lower conductivity (darker zones). This demonstrates the impact of microstructural uniformity on the functional performance of the material.

\section{Conclusion} \label{Sec:conclusion}
This study presents an integrated computational methodology that synergizes machine learning with the shape space theory to advance Cu-Cr-Zr alloy characterization. 

Three key advancements are demonstrated. First, a machine learning framework integrating Feature Augmentation on Geodesic Curves (FAGC) achieves precise predictions for Cu-Cr-Zr alloy properties under data-limited conditions, establishing new benchmarks for small-sample learning in materials science. Second, systematic comparative analysis reveals the decision tree algorithm achieves optimum performance with 100 augmented features, providing quantitative guidance for model selection and feature engineering in microstructure-property mapping. Finally, spatial correlation analysis identifies critical microstructure-conductivity relationships: simplified grain/phase boundary configurations enhance electron transport, while complex interfacial networks induce conductivity reduction.

This work advances academic understanding of microstructure–property relationships while providing industry-reliable solutions for high-performance alloy design under data-scarce conditions. 
Although this work demonstrates the effectiveness of FAGC on experimentally collected microstructural images and data, its robustness and generalizability across a wider range of material image datasets remain to be evaluated. Further investigations using more experimental datasets can provide additional confirmation of the universality and effectiveness of the proposed approach.

\section{Acknowledgement}
This research is sponsored by National Natural Science Foundation of China (Grant No. 52273228).



\bibliographystyle{unsrt}

\bibliography{cas-refs}



\end{document}